\documentclass[twocolumn,10pt]{article}

\pdfoutput=1

\usepackage{docmute}

\usepackage{abstract}
\usepackage{afterpage}
\usepackage{amsmath}
\usepackage{amsthm}
\usepackage{bibunits}
\usepackage{booktabs}
\usepackage[font=small,labelfont=sc]{caption}
\usepackage{chngcntr}
\usepackage{etoolbox}
\usepackage{float}
\usepackage[T1]{fontenc}
\usepackage[margin=1in]{geometry}
\usepackage{graphicx}
\usepackage[colorlinks=true,
            linkcolor=blue,
            citecolor=blue,
            urlcolor=black]{hyperref}
\usepackage{lmodern}
\usepackage{makecell}
\usepackage{multirow}
\usepackage[square]{natbib}
\usepackage{pdflscape}
\usepackage[detect-all]{siunitx}
\usepackage{subcaption}
\usepackage{tabularx}
\usepackage[table]{xcolor}
\usepackage{titling}
\usepackage{xr}

\defaultbibliography{graphconv}
\defaultbibliographystyle{plainnat}

\robustify\bfseries
\definecolor{lightgray}{gray}{0.9}

\newtheorem{property}{Property}

\title{Molecular Graph Convolutions: Moving Beyond Fingerprints}

\author{
    \textbf{\normalsize Steven Kearnes} \\
    \normalsize Stanford University \\
    \texttt{\normalsize kearnes@stanford.edu} \and
    \textbf{\normalsize Kevin McCloskey} \\
    \normalsize Google Inc. \\
    \texttt{\normalsize mccloskey@google.com} \and
    \textbf{\normalsize Marc Berndl} \\
    \normalsize Google Inc. \\
    \texttt{\normalsize marcberndl@google.com} \and
    \textbf{\normalsize Vijay Pande} \\
    \normalsize Stanford University \\
    \texttt{\normalsize pande@stanford.edu} \and
    \textbf{\normalsize Patrick Riley} \\
    \normalsize Google Inc. \\
    \texttt{\normalsize pfr@google.com}
}

\date{}  % Don't put the date in the title.

\let\oldtabular\tabular
\renewcommand{\tabular}{\small\oldtabular}

\begin{document}

\maketitle
\begin{bibunit}
%auto-ignore

%%%%%%%%%%%%%%%%%%%%%%%%%%%%%%%%%%%%%%%%%%%%%%%%%%%%%%%%%%%%%%%%%%%%%%%%%%%%%%%
\begin{abstract}

Molecular ``fingerprints'' encoding structural information are the workhorse of
cheminformatics and machine learning in drug discovery applications. However,
fingerprint representations necessarily emphasize particular aspects of the
molecular structure while ignoring others, rather than allowing the model to
make data-driven decisions. We describe molecular \emph{graph convolutions}, a
machine learning architecture for learning from undirected
graphs, specifically small molecules. Graph convolutions use a simple encoding
of the molecular graph---atoms, bonds, distances, etc.---which allows the model to
take greater advantage of information in the graph structure.
Although graph convolutions do not outperform all fingerprint-based methods,
they (along with other graph-based methods) represent a new paradigm in
ligand-based virtual screening with exciting opportunities for future
improvement.

%\keywords{machine learning \and virtual screening}

\end{abstract}

%%%%%%%%%%%%%%%%%%%%%%%%%%%%%%%%%%%%%%%%%%%%%%%%%%%%%%%%%%%%%%%%%%%%%%%%%%%%%%%
\section{Introduction}\label{sec:intro}

Computer-aided drug design requires representations of molecules that can be
related to biological activity or other experimental endpoints. These
representations encode structural features, physical properties, or activity in
other assays~\citep{todeschini2009molecular, petrone2012rethinking}. The recent
advent of ``deep learning'' has enabled the use of very raw representations that
are less application-specific when building machine learning
models~\citep{lecun2015deep}. For instance, image recognition models that were
once based on complex features extracted from images are now trained exclusively
on the pixels themselves---deep architectures can ``learn'' appropriate
representations for input data. Consequently, deep learning systems for drug
screening or design should benefit from molecular representations that are as
complete and general as possible rather than relying on application-specific
features or encodings.

\begin{figure}[tb]
  % Generated with obabel 2.3.90 using SMILES CC(C)CC1=CC=C(C=C1)C(C)C(=O)O.
  \centering
  \includegraphics[clip,trim=30 125 30 130,width=\linewidth]{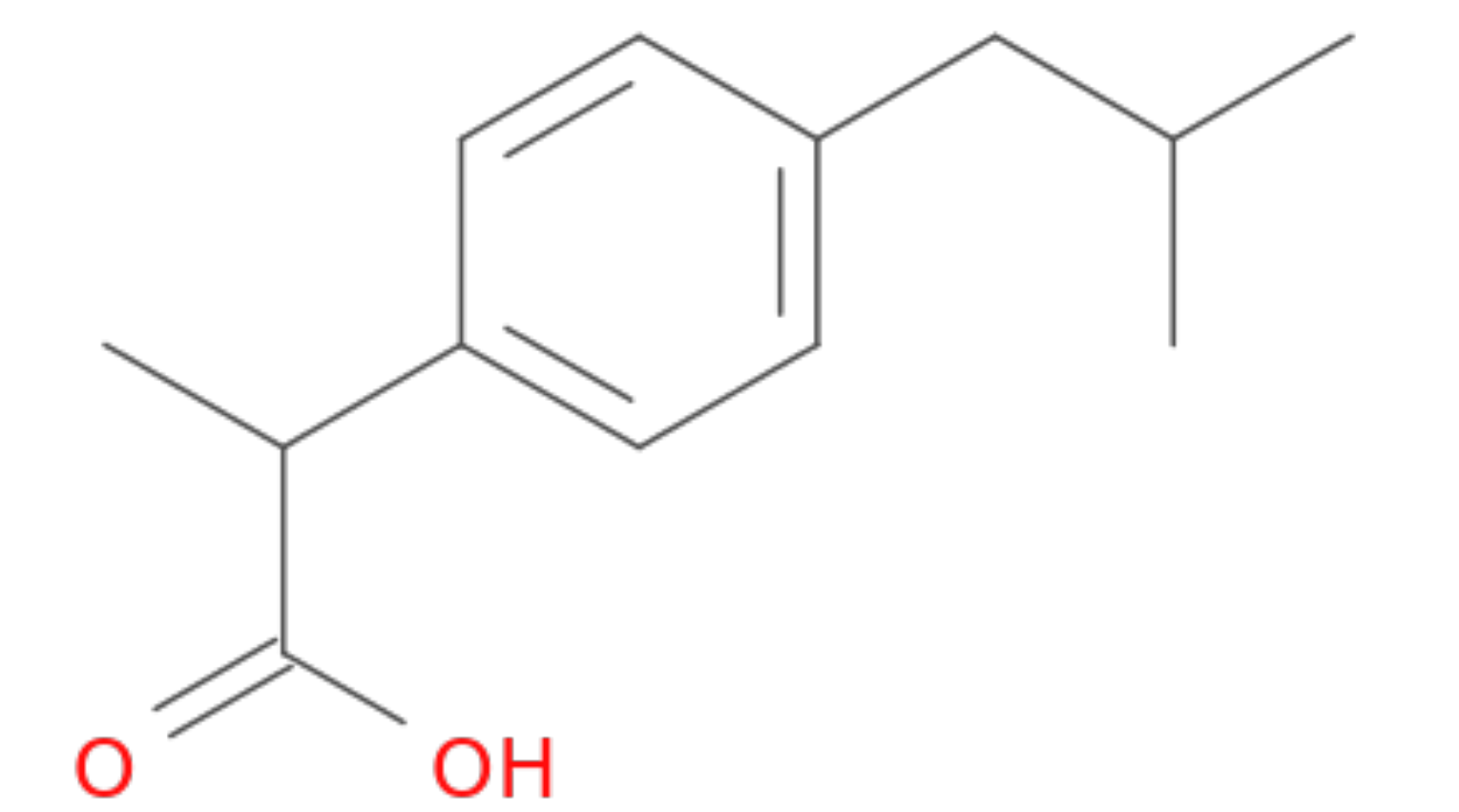}
  \caption{
    Molecular graph for ibuprofen. Unmarked vertices represent carbon atoms, and
    bond order is indicated by the number of lines used for each edge.
  }
  \label{fig:ibuprofen}
\end{figure}

First-year chemistry students quickly become familiar with a common
representation for small molecules: the molecular graph.
\figurename~\ref{fig:ibuprofen} gives an example of the molecular graph for
ibuprofen, an over-the-counter non-steroidal anti-inflammatory drug. The
atoms and bonds between atoms form the nodes and edges, respectively, of the
graph. Both atoms and bonds have associated properties, such as atom type
and bond order. Although the basic molecular graph representation does not
capture the quantum mechanical structure of molecules or necessarily express all
of the information that it might suggest to an expert medicinal chemist, its
ubiquity in academia and industry makes it a desirable starting point for
machine learning on chemical information.

Here we describe molecular \emph{graph convolutions}, a deep learning
system using a representation of small molecules as undirected graphs of atoms.
Graph convolutions extract meaningful features from simple
descriptions of the graph structure---atom and bond properties, and
graph distances---to form molecule-level representations that can be
used in place of fingerprint descriptors in conventional
machine learning applications.

%%%%%%%%%%%%%%%%%%%%%%%%%%%%%%%%%%%%%%%%%%%%%%%%%%%%%%%%%%%%%%%%%%%%%%%%%%%%%%%
\section{Related Work}\label{sec:related_work}

The history of molecular representation is extremely diverse
\citep{todeschini2009molecular} and a full review is outside the scope of this
report. Below we describe examples from several major branches of the field to
provide context for our work. Additionally, we review several recent examples of
graph-centric approaches in cheminformatics.

Much of cheminformatics is based on so-called ``2D'' molecular descriptors that
attempt to capture relevant structural features derived from the molecular
graph. In general, 2D features are computationally inexpensive and easy to
interpret and visualize. One of the most common representations in this class is
extended-connectivity fingerprints (ECFP), also referred to as circular or
Morgan fingerprints~\citep{rogers2010extended}. Starting at each heavy atom, a
``bag of fragments'' is constructed by iteratively expanding outward along bonds
(usually the algorithm is terminated after $2$--$3$ steps). Each unique fragment
is assigned an integer identifier, which is often hashed into a fixed-length
representation or ``fingerprint''. Additional descriptors in this class include
decompositions of the molecular graph into subtrees or fixed-length
paths~\citep{oegraphsimtk}, as well as atom pair (AP) descriptors that encode
atom types and graph distances (number of intervening bonds) for all pairs of
atoms in a molecule~\citep{carhart1985atom}.

Many representations encode 3D information, with special emphasis on molecular
shape and electrostatics as primary drivers of interactions in
real-world systems. For example, rapid overlay of chemical structures
(ROCS) aligns pairs of pre-generated conformers and calculates shape and
chemical (``color'') similarity using Gaussian representations of atoms and
color features defined by a simple force field~\citep{hawkins2007comparison}. ROCS can
also be used to generate alignments for calculation of electrostatic field
similarity~\citep{muchmore2006use}. Ultrafast shape recognition (USR) calculates
alignment-free 3D similarity by comparing distributions of intramolecular
distances~\citep{ballester2007ultrafast}.

The Merck Molecular Activity Challenge~\citep{dahl2012deep} catalyzed interest
in deep neural networks trained on fingerprints and other molecular descriptors.
In particular, multitask neural networks have produced consistent gains relative
to baseline models such as random forest and logistic
regression~\citep{dahl2014multi, ma2015deep, mayr2015deeptox,
ramsundar2015massively}.

Other approaches from both the cheminformatics and the machine learning
community directly operate on graphs in a way similar to how we do here. The ``molecular
graph networks'' of \citet{merkwirth2005automatic} iteratively update a state
variable on each atom with learned weights specific to each atom type--bond type
pair. Similarly, \citet{micheli2009neural} presents a more
general formulation of the same concept of iterated local information transfer
across edges and applies this method to predicting the boiling point of alkanes.

\citet{scarselli2009graph}~similarly defines a local operation on the graph.
They demonstrate that a fixed point
across all the local functions can be found and calculate fixed point solutions
for graph nodes as part of each training step. In another
vein,~\citet{lusci2013deep} convert undirected molecular graphs to a directed
recursive neural net and take an ensemble over multiple conversions.

Recently, \citet{duvenaud2015convolutional} presented an architecture
trying to accomplish many of the same goals as this work. The architecture was
based on generalizing the fingerprint computation such that it can be learned
via backpropagation. They demonstrate that this architecture improves
predictions of solubility and photovoltaic efficiency but not binding affinity.

\citet{bruna2013spectral}~introduce convolutional deep networks on spectral
representations of graphs. However, these methods apply when the graph structure
is fixed across examples and only the labeling/features on individual nodes
varies.

Convolutional networks on non-Euclidean manifolds were described
by~\citet{masci2015geodesic}. The problem addressed was to describe
the shape of the manifold (such as the surface of a human being) in such a way
that the shape descriptor of a particular point was invariant to perturbations
such as movement and deformation. They also describe an approach for combining
local shape descriptors into a global descriptor and demonstrate its use in a
shape classification task.

%%%%%%%%%%%%%%%%%%%%%%%%%%%%%%%%%%%%%%%%%%%%%%%%%%%%%%%%%%%%%%%%%%%%%%%%%%%%%%%
\section{Methods}

\subsection{Deep neural networks}\label{sec:neural_nets}

Neural networks are directed graphs of simulated ``neurons''. Each neuron has a
set of inputs and computes an output. The neurons in early neural nets were
inspired by biological neurons and computed an affine combination of the inputs
followed by a non-linear activation function. Mathematically, if the inputs are
$x_1 \dots x_N$, weights $w_1 \dots w_N$ and bias $b$ are parameters, and $f$ is
the activation function, the output is
\begin{equation}\label{eq:simple_nueron}
  f(b + \sum_i w_i x_i)
\end{equation}
Popular activation functions include the sigmoid function ($f(z) = \frac{1}{1+e^{-z}}$)
and rectified linear unit (ReLU) ($f(z)=0$~if~$z\leq 0$~else~$z$).

Any mostly differentiable function can be
used as the unit of computation for a neuron and in recent years, many other
functions have appeared in published networks, including max and sum.

\emph{Convolution} in neural networks refers to using the same parameters (such as the
$w_i$ in Equation~\ref{eq:simple_nueron}) for different neurons that are attached
to different parts of the input (or previous neurons). In this way, the same
operation is computed for many different subsets of the input.

At the ``top'' of the neural network you have node(s) whose output is the value
you are trying to predict (e.g. the probability that this molecule binds to a
target or the binding affinity). Many output nodes for
different tasks can be added and this is commonly
done~\citep{ma2015deep, ramsundar2015massively}. In this way, different
output tasks can share the computation and model parameters in lower parts of
the network before using their own parameters for the final output steps.

The \emph{architecture} of a neural network refers to the choice of the number of
neurons, the type of computation each one does (including what learnable
parameters they have), which parameters are shared across neurons, and how the
output of one neuron is connected to the input of another.

In order to train the network, you first have to choose a \emph{loss function}
describing the penalty for the network producing a set of outputs which differ
from the outputs in the training example. For example, for regression
problems, the L2 distance between the predicted and actual values is commonly
used. The objective of training is then to find a set of parameters for the
network that minimizes the loss function. Training is done with the well known
technique of back-propagation~\citep{rumelhart:backprop} and stochastic gradient
descent.

\subsection{Desired invariants of a model}\label{sec:architecture}

A primary goal of designing a deep learning architecture is to restrict the set
of functions that can be learned to ones that match the desired properties from
the domain. For example, in image understanding, spatial convolutions force the
model to learn functions that are invariant to translation.

For a deep learning architecture taking a molecular graph as input, some
arbitrary choice must be made for the order that the various atoms and bonds
are presented to the model. Since that choice is arbitrary, we want:
\begin{property}[Order invariance]
  \label{prop:order_invariance}
  The output of the model should be invariant to the order that the atom and
  bond information is encoded in the input.
\end{property}

Note that many current procedures for fingerprinting molecules achieve
Property~\ref{prop:order_invariance}. We will now gradually construct an
architecture which achieves Property~\ref{prop:order_invariance} while
making available a richer space of learnable parameters.

The first basic unit of representation is an \emph{atom layer} which
contains an $n$-dimensional vector associated with each atom. Therefore the
atom layer is a 2~dimensional matrix indexed first by atom. Part of the original
input will be encoded in such an atom layer and the details of how we construct
the original input vector are discussed in Section~\ref{sec:input_features}. The
next basic unit of representation is a \emph{pair layer} which contains an
$n$-dimensional vector associated with each pair of atoms. Therefore, the pair
layer is a 3~dimensional matrix where the first two dimensions are indexed by
atom. Note that the pair input can contain information not just about edges but
about any arbitrary pair. Notably, we will encode the graph distance (length of
shortest path from one atom to the other) in the input pair layer. The order of
the atom indexing for the atom and pair layer inputs must be the same.

We will describe various operations to compute new atom and pair layers with
learnable parameters at every step. Notationally, let $A^x$ be the value of a
particular atom layer $x$ and $P^y$ be the value of a particular pair layer
$y$. The inputs that produce those values should be clear from the
context. $A^x_a$ refers to the value of atom $a$ in atom layer $x$ and
$P^y_{(a,b)}$ refers to the value of pair $(a,b)$ in pair layer $y$.

In order to achieve Property~\ref{prop:order_invariance} for the overall
architecture, we need a different type of invariance for each atom and pair
layer.

\begin{property}[Atom and pair permutation invariance]
  \label{prop:atom_pair_permutation}
  The values of an atom layer and pair permute with the original input layer
  order. More precisely, if the inputs are permuted with a permutation operator
  $Q$, then for all layers $x, y$, $A^x$ and $P^y$ are permuted with operator
  $Q$ as well.
\end{property}

In other words, Property~\ref{prop:atom_pair_permutation} means that from a
single atom's (or pair's) perspective, its value in every layer is invariant to
the order of the other atoms (or pairs).

Since molecules are undirected graphs, we will also maintain the following:
\begin{property}[Pair order invariance]
  \label{prop:pair_order}
  For all pair layers $y$, $P^y_{(a,b)} = P^y_{(b,a)}$
\end{property}
Property~\ref{prop:pair_order} is easy to achieve at the input layer and the
operations below will maintain this.

Properties~\ref{prop:atom_pair_permutation} and~\ref{prop:pair_order} make it
easy to construct a molecule-level representation from an atom or pair such that
the molecule-level representation achieves
Property~\ref{prop:order_invariance} (see Section~\ref{sec:features}).

\subsection{Invariant-preserving operations}\label{sec:operations}

We now define a series of operations that maintain the above properties.

Throughout, $f$ represents an arbitrary function and $g$ represents an arbitrary
\emph{commutative} function ($g$ returns the same result regardless of the
order the arguments are presented). In this work, $f$ is a learned linear
operator with a rectified linear (ReLU) activation function and $g$ is a sum.

The most trivial operation is to combine one or more layers of the same type
by applying the same operation to every atom or pair. Precisely, this
means if you have layers $x1, x2, \dotsc, xn$ and function $f$, you can compute
a new atom layer from the previous atom layer ($A\rightarrow A$) as
\begin{equation}
  A^y_a = f(A^{x1}_a, A^{x2}_a, \dotsc, A^{xn}_a)
\end{equation}
or pair layer from the previous pair layer ($P\rightarrow P$) as
\begin{equation}
  P^y_{a,b} = f(P^{x1}_{a,b}, P^{x2}_{a,b}, \dotsc, P^{xn}_{a,b})
\end{equation}
Since we apply the same function for every atom/pair, we refer to this as a
convolution. All the transformations we develop below will have this convolution
nature of applying the same operation to every atom/pair, maintaining
Property~\ref{prop:atom_pair_permutation}.

When operating on pairs of atoms, instead of putting all pairs through
this function, you could select a subset. In
Section~\ref{sec:neighbors} we show experiments for restricting the
set of pairs to those that are less than some graph distance away.

\begin{figure}[tb]
  % From
  % https://docs.google.com/drawings/d/1vdlGkswGOC6uzomum0pmrzrcTkXhaZa5xg2r5KVlGVo/edit
  \centering
  \includegraphics[clip,trim=0 360 525 0]{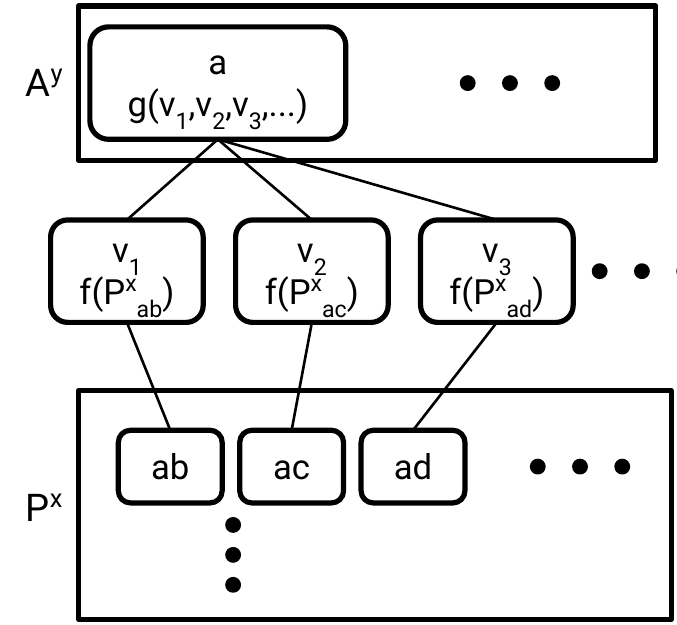}
  \caption{$P \rightarrow A$ operation. $P^x$ is a matrix containing features
  for atom pairs $ab$, $ac$, $ad$, etc. The $v_i$ are intermediate values
  obtained by applying $f$ to features for a given atom pair. Applying $g$ to
  the intermediate representations for all atom pairs involving a given atom
  (e.g.~$a$) results in a new atom feature vector for that atom.}
  \label{fig:p-to-a}
\end{figure}

Next, consider an operation that takes a pair layer $x$ and constructs an atom
layer $y$ ($P \rightarrow A$). The operation is depicted in
\figurename~\ref{fig:p-to-a}. Formally:
\begin{equation}
    A^y_a = g(f(P^x_{(a,b)}), f(P^x_{(a,c)}), f(P^x_{(a,d)}), ...)
\end{equation}
In other words, take all pairs of which $a$ is a part, run them through $f$, and
combine them with $g$. Note that Property~\ref{prop:pair_order} means we can
choose an arbitrary one of $P^x_{(a,b)}$ or $P^x_{(b,a)}$.

\begin{figure}[tb]
  % From
  % https://docs.google.com/drawings/d/1vdlGkswGOC6uzomum0pmrzrcTkXhaZa5xg2r5KVlGVo/edit
  \centering
  \includegraphics[clip,trim=0 375 520 0]{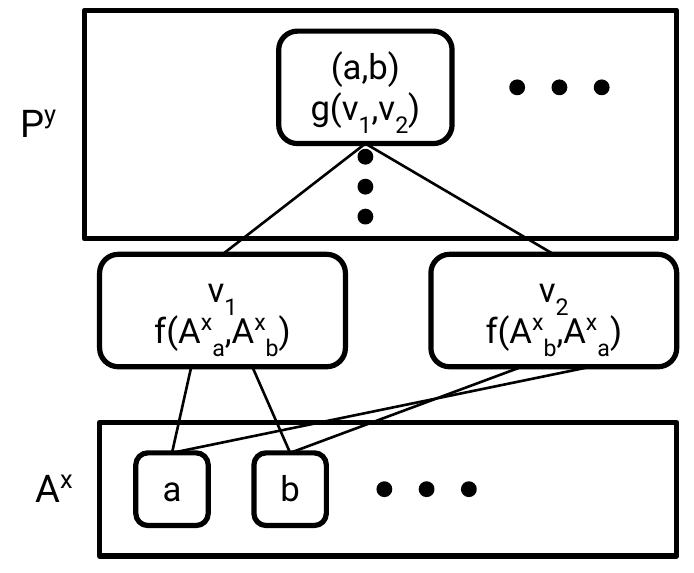}
  \caption{$A \rightarrow P$ operation. $A^x$ is a matrix containing features
  for atoms $a$, $b$, etc. The $v_i$ are intermediate values obtained by
  applying $f$ to features for a given pair of atoms concatenated in both
  possible orderings ($ab$ and $ba$). Applying $g$ to these intermediate ordered
  pair features results in an order-independent feature vector for atom pair
  $ab$.}
  \label{fig:a-to-p}
\end{figure}
The most interesting construction is making a pair layer from an atom
layer ($A \rightarrow P$). The operation is graphically depicted in
\figurename~\ref{fig:a-to-p} and formally as
\begin{equation}
    P^y_{ab} = g(f(A^x_a, A^x_b), f(A^x_b, A^x_a))
\end{equation}
Note that just applying $g$ to $A^x_a$ and $A^x_b$ would maintain
Properties~\ref{prop:atom_pair_permutation} and~\ref{prop:pair_order} but we use
this more complex form. While commutative operators (such as max pooling) are
common in neural networks, commutative operators \emph{with learnable
parameters} are not common. Therefore, we use $f$ to give learnable parameters
while maintaining the desired properties.

\begin{figure}[tb]
  % From
  % https://docs.google.com/drawings/d/14VWVx6SUsQ2mkhYf00iLBP6g_xw0MoG4R588LvYetaw/edit
  \centering
  \includegraphics[clip,trim=0 365 535 25]{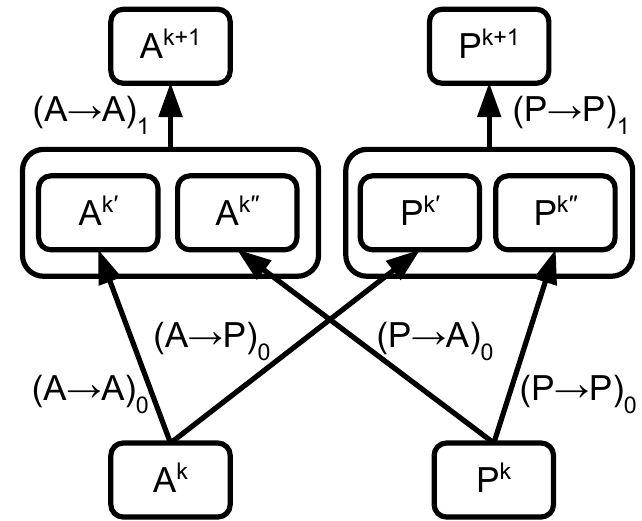}
  \caption{Weave module. This module takes matrices $A^k$ and $P^k$ (containing
  atom and pair features, respectively) and combines $A\rightarrow A$,
  $P\rightarrow P$, $P\rightarrow A$, and $A\rightarrow P$ operations to yield a
  new set of atom and pair features ($A^{k+1}$ and $P^{k+1}$, respectively). The
  output atom and pair features can be used as input to a subsequent Weave
  module, which allows these modules to be stacked in series to an arbitrary
  depth.}
  \label{fig:weave}
\end{figure}
Once we have all the primitive operations on atom and pair layers
($A\rightarrow A$, $P\rightarrow P$, $P\rightarrow A$, $A\rightarrow P$), we
can combine these into one module.  We call this the Weave module
(\figurename~\ref{fig:weave}) because the atoms and pair layers cross back and forth
to each other. The module can be stacked to an arbitrary depth similar to the
Inception module that inspired it~\citep{szegedy2014going}.
Deep neural networks with many layers (e.g. for computer vision) learn
progressively more general features---combinations of lower-level features---in
a hierarchical manner~\citep{lecun2015deep}. By analogy, successive Weave
modules can produce more informative representations of the original input.
Additionally, stacked Weave modules with limited maximum atom pair distance
progressively incorporate longer-range information at each layer.

\subsection{Molecule-level features}
\label{sec:features}

The construction of the Weave module maintains
Properties~\ref{prop:atom_pair_permutation} and~\ref{prop:pair_order}. What
about overall order invariance (Property~\ref{prop:order_invariance})? At the
end of a stack of Weave modules we are left with an $n$-dimensional
vector associated with every atom and an $m$-dimensional vector associated with
every pair. We need to turn this into a molecule-level representation with some
commutative function of these vectors.

In related work~\citep{merkwirth2005automatic,
duvenaud2015convolutional, lusci2013deep}, a simple unweighted sum is
often used to combine order-dependent atom features into order-independent
molecule-level features. However, reduction to a single value does not capture
the distribution of learned features. We experimented with an alternative
approach and created ``fuzzy'' histograms for each dimension of the feature
vector.

A fuzzy histogram is described by a set of \emph{membership functions} that are
functions with range $[0, 1]$ representing the membership of the point in each
histogram bin~\citep{zadeh1965fuzzy}. A standard histogram has membership functions which
are 1 in the bin and 0 everywhere else. For each point, we normalize so that the
total contribution to all bins is 1. The value of a bin in the histogram over
all points is just the sum of the normalized contributions for all the points.

\begin{figure}[tb]
  % From code/fuzzy_histogram.py
  \centering
  \includegraphics[width=\linewidth]{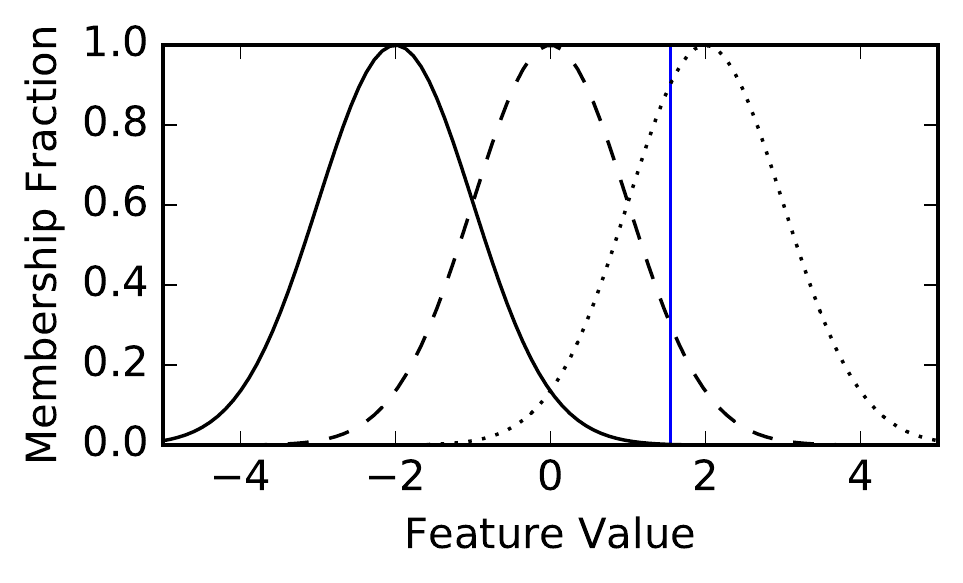}
  \caption{
    Fuzzy histogram with three Gaussian ``bins''. Each curve represents the
    membership function for a different bin, indicating the degree to which a
    point contributes to that bin. The vertical blue line represents an example
    point which contributes normalized densities of $<0.01$, $\sim0.25$, and
    $\sim0.75$ to the bins (from left to right).
  }
  \label{fig:fuzzy_histogram}
\end{figure}

\figurename~\ref{fig:fuzzy_histogram} gives an example of a fuzzy histogram
composed of three Gaussian bins. A histogram is constructed for each dimension
of the feature vectors and the concatenation of those histograms is the
molecule-level representation.

In this work we used Gaussian membership functions (which are unnormalized
versions of the standard Gaussian PDF) with eleven bins spanning a Gaussian
distribution with mean of zero and unit standard deviation, shown in
\figurename~\ref{appendix:fig:gaussian_bins}. These bins were chosen somewhat arbitrarily to cover the
expected distribution of incoming features and were not optimized further (note
that the incoming features were batch normalized; see
Section~\ref{sec:training}).

Throughout this paper, we construct the molecule-level features only from the
top-level atom features and not the pair features. This is to restrict the total
number of feature vectors that must be summarized while still providing
information about the entire molecule. Note, however, that the initial and
intermediate pair features can influence the final atom features through Weave
module operations.

Before the molecule-level featurization, we do one final convolution on the
atoms. Since molecule-level featurization can be a major bottleneck in the
model, this convolution expands the depth so that each dimension of the atom
feature vector contains less information and therefore less information is lost
during the molecule-level featurization. On this convolution, we do not use a
ReLU activation function to avoid the histogram having many points at zero.

Once you have a molecule-level representation, this becomes a more standard
multitask problem. We follow the common
approach~\citep{ramsundar2015massively, ma2015deep, mayr2015deeptox}
of a small number of fully connected layers on top of the molecule-level
features followed by standard softmax classification.

\begin{figure}[tb]
  % From
  % https://docs.google.com/drawings/d/1SJ_2i4CJR4f5Co2Cuvfh68o9lQ2Mof148LtKsyzXjgE/edit
  \centering
  \includegraphics[clip,trim=0 270 545 0,width=0.75\linewidth]{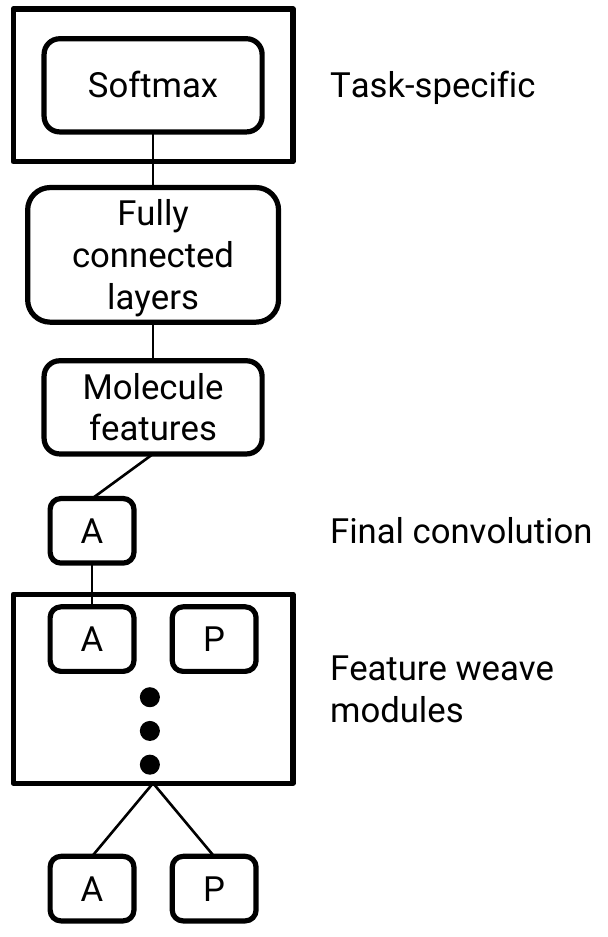}
  \caption{Abstract graph convolution architecture. In the current
  implementation, only the final atom features are used to generate
  molecule-level features.}
  \label{fig:abstract_architecture}
\end{figure}

The overall architecture is depicted in
\figurename~\ref{fig:abstract_architecture}.
\tablename~\ref{table:hyperparameters} lists hyperparameters and default values
for graph convolution models.
In models with multiple Weave modules it is conceivable to vary the
convolution depths in a module-specific way. However, the models in this work
used the same settings for all Weave modules.

Our current implementation imposes an upper limit on the number of heavy atoms
represented in the initial featurization. For molecules that have more than the
maximum number of atoms, only a subset of atoms (and therefore atom pairs) are
represented in the input encoding. This subset depends on the order in which the
atoms are traversed by the featurization code and should be considered
arbitrary. In this work we set the maximum number of atoms to 60, and only
\num{814} of the \num{1442713} unique molecules in our datasets (see
Section~\ref{sec:graphconv_datasets}) exceed this limit.

\begin{table*}[htbp]
  \caption{Graph convolution model hyperparameters.}
  \label{table:hyperparameters}
  \centering
  \rowcolors{1}{}{lightgray}
  \begin{tabular}{ l l r }
  \toprule
  Group & Hyperparameter & Default Value \\
  \midrule
  \cellcolor{white} & Maximum number of atoms per molecule & 60 \\
  \multirow{-2}{*}{\cellcolor{white}Input} &
      Maximum atom pair graph distance & 2 \\
  \midrule
  \cellcolor{white} & Number of Weave modules & 1 \\
  \cellcolor{white} & $(A \rightarrow A)_0$ convolution depth & 50 \\
  \cellcolor{white} & $(A \rightarrow P)_0$ convolution depth & 50 \\
  \cellcolor{white} & $(P \rightarrow P)_0$ convolution depth & 50 \\
  \cellcolor{white} & $(P \rightarrow A)_0$ convolution depth & 50 \\
  \cellcolor{white} & $(A \rightarrow A)_1$ convolution depth & 50 \\
  \multirow{-7}{*}{\cellcolor{white}Weave} & $(P \rightarrow P)_1$ convolution depth & 50 \\
  \midrule
  \cellcolor{white} & Final atom layer convolution depth & 128 \\
  \multirow{-2}{*}{\cellcolor{white}Reduction} &
      Reduction to molecule-level features & Gaussian histogram \\
  \midrule
  \cellcolor{white}Post-reduction &
      Fully-connected layers (number of units per layer) & 2000, 100 \\
  \midrule
  \cellcolor{white} & Batch size & 96 \\
  \cellcolor{white} & Learning rate & 0.003 \\
  \multirow{-3}{*}{\cellcolor{white}Training} & Optimization method & Adagrad \\
  \bottomrule
  \end{tabular}
\end{table*}

\subsection{Input featurization}\label{sec:input_features}

\begin{table*}[htbp]
    \caption{Atom features.}
    \label{table:atom_features}
    \centering
    \rowcolors{1}{}{lightgray}
    \begin{tabular}{ l l r }
    \toprule
    Feature & Description & Size \\
    \midrule
    Atom type$^*$ & H, C, N, O, F, P, S, Cl, Br, I, or metal (one-hot or null).
                  & $11$ \\
    Chirality & R or S (one-hot or null). & $2$ \\
    Formal charge & Integer electronic charge. & $1$ \\
    Partial charge & Calculated partial charge. & $1$ \\
    Ring sizes & For each ring size ($3$--$8$), the number of rings that include
                 this atom. & $6$ \\
    Hybridization & sp, sp$^2$, or sp$^3$ (one-hot or null). & $3$ \\
    Hydrogen bonding & Whether this atom is a hydrogen bond donor and/or
                       acceptor (binary values). & $2$ \\
    Aromaticity & Whether this atom is part of an aromatic system. & $1$ \\
    \midrule
    \cellcolor{white} & \cellcolor{white} & \cellcolor{white}$27$ \\
    \bottomrule
    \multicolumn{3}{l}{\small * Included in the ``simple'' featurization (see
                       Section~\ref{sec:simple_features}).}
    \end{tabular}
\end{table*}

\begin{table*}[htbp]
    \caption{Atom pair features.}
    \label{table:pair_features}
    \centering
    \rowcolors{1}{}{lightgray}
    \begin{tabular}{ l l r }
    \toprule
    Feature & Description & Size \\
    \midrule
    Bond type$^*$ & Single, double, triple, or aromatic (one-hot or null).
                  & $4$ \\
    Graph distance$^*$ &
        \makecell[l]{For each distance ($1$--$7$), whether the shortest path \\
        between the atoms in the pair is less than or equal \\
        to that number of bonds (binary values).}
        & $7$ \\
    Same ring & Whether the atoms in the pair are in the same ring. & $1$ \\
    \midrule
     & & $12$ \\
    \bottomrule
    \multicolumn{3}{l}{\cellcolor{white}\small * Included in the ``simple''
                       featurization (see Section~\ref{sec:simple_features}).}
    \end{tabular}
\end{table*}

The initial atom and pair features are summarized in
\tablename~\ref{table:atom_features} and \tablename~\ref{table:pair_features},
respectively. The features are a mix of floating point, integer, and binary
values (all encoded as floating point numbers in the network). The feature set
is intended to be broad, but not necessarily exhaustive, and we recognize that
some features can potentially be derived from or correlated to a subset of the
others (e.g. atom hybridization can be determined by inspecting the
bonds that atom makes). We performed experiments using a ``simple'' subset of
these features in an effort to understand their relative contributions to
learning (Section~\ref{sec:simple_features}), but many other questions about
specifics of the input featurization are left to future work.

All features were generated with RDKit \citep{landrumrdkit}, including Gasteiger
atomic partial charges~\citep{gasteiger1980iterative}. Although our
featurization includes space for hydrogen atoms, we did not use explicit
hydrogens in any of our experiments in order to conserve memory and emphasize
contributions from heavy atoms.

Other deep learning applications with more ``natural'' inputs such as computer
vision and speech recognition still require some input engineering; for example,
adjusting images to a specific size or scale, or transforming audio into the
frequency domain.  Likewise, the initial values for the atom and pair layers
describe these primitives in terms of properties that are often considered by
medicinal chemists and other experts in the field, allowing the network to use
or ignore them as needed for the task at hand. One of the purposes of this work
is to demonstrate that learning can occur with as little preprocessing as
possible. Accordingly, we favor simple descriptors that are more or less
``obvious''.

\subsection{Datasets}
\label{sec:graphconv_datasets}

We used a dataset collection nearly identical to the one described
by~\citet{ramsundar2015massively} except for some changes to the data
processing pipeline (including the duplicate merging process for the Tox21
dataset) and different cross-validation fold divisions. Briefly, there are 259
datasets divided into four groups indicating their source: PubChem BioAssay
\citep{wang2012pubchem} (PCBA, 128 datasets), the ``maximum unbiased
validation''
datasets constructed by Rohrer and Baumann~\citep{rohrer2009maximum} (MUV, 17
datasets), the enhanced directory of useful decoys~\citep{mysinger2012directory}
(DUD-E, 102 datasets), and the training set for the Tox21
challenge~(see~\citet{mayr2015deeptox}) (Tox21, 12 datasets). The combined
dataset
contained over 38~M data points and included targets from many different
biological classes.

\subsection{Model training and evaluation}
\label{sec:training}

Graph convolution and traditional neural network models were implemented with
TensorFlow~\citep{abaditensorflow}, an open-source library for machine learning.
Models were evaluated by the area under the receiver operating characteristic
curve (ROC AUC, or simply AUC) as recommended by \citet{jain2008recommendations}. We used 5-fold
stratified cross-validation, where each fold-specific model used 60\% of the
data for training, 20\% for validation (early stopping/model selection), and 20\% as a test set.

Graph convolution models were trained for 10--20~M steps using the Adagrad
optimizer~\citep{duchi2011adaptive} with learning rate 0.003 and batch size 96,
with periodic checkpointing. All convolution and fully-connected layer outputs
were batch normalized~\citep{ioffe2015batch} prior to applying the ReLU
nonlinearity. Training was parallelized over 96 CPUs (or 96 GPUs in the case of
the W$_4$N$_2$ model) and required several days for each model. Adding
additional Weave modules significantly increased training time. However, models
trained on smaller datasets (see Section~\ref{sec:other_comparisons}) trained
much faster.

To establish a baseline, we also trained pyramidal (2000, 100) multitask neural
network (PMTNN)~\citep{ramsundar2015massively}, random forest (RF), and logistic
regression (LR) models using Morgan fingerprints with radius 2 (essentially
equivalent to ECFP4) generated with RDKit~\citep{landrumrdkit}. As a very simple
baseline, we also computed Tanimoto similarity to all training set actives and
used the maximum similarity score as the active class probability (MaxSim).

The PMTNN had two hidden layers (with 2000 and 100 units, respectively) with
rectified linear activations, and each fold-specific model was trained for
40--50~M steps using the SGD optimizer with batch size 128 and a learning rate
of 0.0003, with periodic checkpointing. Additionally, this model used 0.25
dropout~\citep{srivastava2014dropout}, initial weight standard deviations of
0.01 and 0.04 and initial biases of 0.5 and 3.0 in the respective hidden layers.
This model did not use batch normalization.

Logistic regression (LR) models were
trained with the \texttt{LogisticRegression} class in
scikit-learn~\citep{pedregosa2011scikit} using the \texttt{`lbfgs'} solver and
a maximum of \num{10000} iterations. Values for the regularization strength
(\texttt{C}) parameter were chosen by grid search, using the held-out validation
set for model selection. Random forest (RF) models were trained using the
scikit-learn \texttt{RandomForestClassifier} with 100 trees.

In graph convolution and PMTNN models, active compounds were weighted in the
cost function such that the total active weight equalled the total inactive
weight within each dataset (logistic regression and random forest models also
used these weights as the \texttt{sample\_weight} argument to their \texttt{fit}
methods). Furthermore, graph convolution and PMTNN models were evaluated in a
task-specific manner by choosing the training checkpoint with the best
validation set AUC for each task. We note that some fold-specific models had a
small number of tasks were not ``converged'' in the sense that their validation
set AUC scores were still increasing when training was halted, and that the
specific tasks that were not converged varied from model to model.

To statistically compare graph convolution and baseline models, we report three
values for each dataset group: (1)~median 5-fold mean AUC over all datasets,
(2)~median difference in per-dataset 5-fold mean AUC ($\Delta$AUC) relative to
the PMTNN baseline, and (3)~a 95\% Wilson score interval for the sign test
statistic relative to the PMTNN baseline. The sign test estimates the
probability that a model will achieve a higher 5-fold mean AUC than the PMTNN
baseline; models with sign test confidence intervals that do not include 0.5
are considered significantly different in their performance (the median
$\Delta$AUC can be used as a measure of effect size). To calculate these
intervals, we used the \texttt{proportion\_confint} function in statsmodels
\citep{seabold2010statsmodels} version 0.6.1 with \texttt{method=`wilson'} and
\texttt{alpha=0.05}, counting only non-zero differences in the sign test.  We do
not report values for the DUD-E dataset group since all models achieved $>0.98$
median 5-fold mean AUC.

As a general note, confidence intervals for box plot medians were computed as
${\pm 1.57 \times \text{IQR} / \sqrt{N}}$ \citep{mcgill1978variations} and do
not necessarily correspond to sign test confidence intervals.

\subsection{Comparisons to other methods}
\label{sec:other_comparisons}

In addition to the baseline models described in Section~\ref{sec:training},
there are many other methods that would be interesting to compare to our graph
convolution models. In particular,~\citet{duvenaud2015convolutional} described
``neural fingerprints'' (NFP), a related graph-based method. The original
publication describing NFP reported mean squared errors (MSE) on datasets for
aqueous solubility, drug efficacy, and photovoltaic efficiency. We trained
multitask graph convolution models on these datasets using 5-fold
cross-validation (note that the published NFP models were single-task).

Additionally, we report results on a dataset used to validate the influence
relevance voter (IRV) method of \citet{swamidass2009influence}, which is a
hybrid of neural networks and $k$-nearest neighbors. The original publication
reported results for two datasets, HIV and DHFR, but the latter was no longer
available from its original source. We trained graph convolution models on the
HIV dataset using 10-fold stratified cross-validation. In each cross-validation
round, one fold each was used for testing and validation (early stopping), and
the remaining folds were used for training. We note that RDKit was only able to
process \num{41476} of the \num{42678} SMILES strings in the HIV dataset.
We report performance on this dataset using both ROC AUC and
BEDROC~\citep{truchon2007evaluating} with $\alpha=20$.

Although we expect our results on these datasets to provide reasonable
comparisons to published data,  differences in fold assignments and variations
in dataset composition due to featurization failures mean that the comparisons
are not perfect.

%%%%%%%%%%%%%%%%%%%%%%%%%%%%%%%%%%%%%%%%%%%%%%%%%%%%%%%%%%%%%%%%%%%%%%%%%%%%%%%
\section{Results}
\label{sec:results}

\subsection{Proof of concept}

With so many hyperparameters to adjust, we sought to establish a centerpoint
from which to investigate specific questions. After several experiments, we
settled on a simple model with two Weave modules, a maximum atom pair distance
of 2, Gaussian histogram molecule-level reductions, and two fully-connected
layers of size 2000 and 100, respectively. Notationally, we refer to this model
as W$_2$N$_2$. \tablename~\ref{table:results} shows the performance of the
W$_2$N$_2$ model and related models derived from this centerpoint by varying a
single hyperparameter. Additionally, \tablename~\ref{table:results} includes
results for several baseline models: MaxSim, logistic regression (LR), random
forest (RF), and pyramidal~(2000,~100) multitask neural network (PMTNN) models
trained on Morgan fingerprints.

Several graph convolution models achieved performance comparable to the baseline
PMTNN on the classification tasks in our dataset collection, which is a
remarkable result considering the simplicity of our input representation. For
example, the centerpoint W$_2$N$_2$ model is statistically indistinguishable
from the PMTNN for the PCBA, MUV, and Tox21 dataset groups (we do not report
results for the DUD-E dataset group because all models achieved extremely high
median AUC scores). Additionally, many of the graph convolution models with
worse performance than the PMTNN (i.e. sign test confidence intervals
excluding 0.5) had very small effective differences as measured by median
$\Delta$AUC.

As an additional measure of model performance, we also calculated ROC
enrichment~\citep{jain2008recommendations} scores at the following false
positive rates: 1\%, 5\%, 10\%, and 20\%. Enrichment scores are reported in
Section~\ref{sec:roc_enrichment} and show that graph convolution models
generally performed worse than or comparable to the PMTNN. We note that the
analysis of model performance and hyperparameter optimization that follows is
based only on ROC AUC scores.

\afterpage{
\begin{landscape}
\vspace*{\fill}  % http://tex.stackexchange.com/questions/97050
\begin{table}[htb]
    \caption{Median 5-fold mean AUC values for reported models. Graph
    convolution models are labeled as W$_x$N$_y$, where $x$ and $y$ denote the
    number of Weave modules and the maximum atom pair distance, respectively
    (see the text for descriptions of the simple, sum, and RMS models). All
    graph convolution models fed into a Pyramidal~(2000,~100) MTNN after the
    molecule-level feature reduction step. MaxSim, logistic regression (LR),
    random forest (RF), and pyramidal~(2000,~100) multitask neural network (PMTNN) baselines used
    Morgan fingerprints as input. For each model, we report the median
    $\Delta$AUC and the 95\% Wilson score interval for a sign test estimating
    the probability that a given model will outperform the PMTNN baseline (see
    Section~\ref{sec:training}). Bold values indicate sign test confidence
    intervals that do not include 0.5.}
    \label{table:results}
    \centering
    \rowcolors{1}{}{lightgray}
    \begin{tabular}{ l S S c S S c S S c }
    \toprule
     & \multicolumn{3}{c}{PCBA $(n=128)$} &
       \multicolumn{3}{c}{MUV $(n=17)$} &
       \multicolumn{3}{c}{Tox21 $(n=12)$} \\
    \cmidrule(lr){2-4} \cmidrule(lr){5-7} \cmidrule(lr){8-10}
    Model & {\makecell{Median \\ AUC}} &
            {\makecell{Median \\ $\Delta$AUC}} &
            {\makecell{Sign Test \\ 95\% CI}} &
            {\makecell{Median \\ AUC}} &
            {\makecell{Median \\ $\Delta$AUC}} &
            {\makecell{Sign Test \\ 95\% CI}} &
            {\makecell{Median \\ AUC}} &
            {\makecell{Median \\ $\Delta$AUC}} &
            {\makecell{Sign Test \\ 95\% CI}} \\
    \midrule
    MaxSim & 0.754 & \bfseries \color{red} -0.137 & \bfseries \color{red} (\num{0.00}, \num{0.04}) & 0.638 & \bfseries \color{red} -0.136 & \bfseries \color{red} (\num{0.01}, \num{0.27}) & 0.728 & \bfseries \color{red} -0.131 & \bfseries \color{red} (\num{0.00}, \num{0.24}) \\
    LR & 0.838 & \bfseries \color{red} -0.059 & \bfseries \color{red} (\num{0.04}, \num{0.13}) & 0.736 & \bfseries \color{red} -0.070 & \bfseries \color{red} (\num{0.10}, \num{0.47}) & 0.789 & \bfseries \color{red} -0.073 & \bfseries \color{red} (\num{0.01}, \num{0.35}) \\
    RF & 0.804 & \bfseries \color{red} -0.092 & \bfseries \color{red} (\num{0.02}, \num{0.10}) & 0.655 & \bfseries \color{red} -0.135 & \bfseries \color{red} (\num{0.01}, \num{0.27}) & 0.802 & \bfseries \color{red} -0.047 & \bfseries \color{red} (\num{0.01}, \num{0.35}) \\
    PMTNN & 0.905 &  &  & 0.869 &  &  & 0.854 &  &  \\
    \midrule
    W$_2$N$_2$-simple & 0.905 & \bfseries \color{red} -0.003 & \bfseries \color{red} (\num{0.27}, \num{0.44}) & 0.849 & 0.012 & (\num{0.36}, \num{0.78}) & 0.866 & 0.003 & (\num{0.39}, \num{0.86}) \\
    W$_2$N$_2$-sum & 0.898 & \bfseries \color{red} -0.011 & \bfseries \color{red} (\num{0.16}, \num{0.31}) & 0.818 & -0.014 & (\num{0.17}, \num{0.59}) & 0.848 & -0.010 & (\num{0.09}, \num{0.53}) \\
    W$_2$N$_2$-RMS & 0.902 & \bfseries \color{red} -0.007 & \bfseries \color{red} (\num{0.20}, \num{0.35}) & 0.851 & -0.026 & (\num{0.13}, \num{0.53}) & 0.854 & \bfseries \color{red} -0.007 & \bfseries \color{red} (\num{0.05}, \num{0.45}) \\
    W$_1$N$_2$ & 0.905 & \bfseries \color{red} -0.007 & \bfseries \color{red} (\num{0.20}, \num{0.35}) & 0.840 & -0.002 & (\num{0.26}, \num{0.69}) & 0.849 & -0.009 & (\num{0.09}, \num{0.53}) \\
    W$_2$N$_1$ & 0.908 & \bfseries \color{red} -0.003 & \bfseries \color{red} (\num{0.30}, \num{0.46}) & 0.858 & -0.016 & (\num{0.17}, \num{0.59}) & 0.867 & -0.002 & (\num{0.19}, \num{0.68}) \\
    W$_2$N$_2$ & 0.909 & 0.000 & (\num{0.42}, \num{0.59}) & 0.847 & -0.004 & (\num{0.22}, \num{0.64}) & 0.862 & 0.004 & (\num{0.32}, \num{0.81}) \\
    W$_2$N$_3$ & 0.906 & -0.001 & (\num{0.38}, \num{0.55}) & 0.838 & -0.013 & (\num{0.26}, \num{0.69}) & 0.861 & 0.000 & (\num{0.25}, \num{0.75}) \\
    W$_2$N$_4$ & 0.908 & -0.001 & (\num{0.37}, \num{0.54}) & 0.836 & -0.008 & (\num{0.17}, \num{0.59}) & 0.858 & 0.001 & (\num{0.39}, \num{0.86}) \\
    W$_2$N$_\infty$ & 0.897 & \bfseries \color{red} -0.008 & \bfseries \color{red} (\num{0.12}, \num{0.25}) & 0.841 & \bfseries \color{red} -0.025 & \bfseries \color{red} (\num{0.10}, \num{0.47}) & 0.846 & -0.006 & (\num{0.14}, \num{0.61}) \\
    W$_3$N$_2$ & 0.906 & 0.000 & (\num{0.44}, \num{0.61}) & 0.875 & 0.010 & (\num{0.31}, \num{0.74}) & 0.859 & 0.004 & (\num{0.47}, \num{0.91}) \\
    W$_4$N$_2$ & 0.907 & -0.001 & (\num{0.33}, \num{0.50}) & 0.856 & -0.007 & (\num{0.22}, \num{0.64}) & 0.862 & 0.004 & (\num{0.32}, \num{0.81}) \\
    \bottomrule
    \end{tabular}
\end{table}
\vspace*{\fill}
\end{landscape}
}

We also trained graph convolution models on some additional datasets in order to
compare to the ``neural fingerprints'' (NFP) of
\citet{duvenaud2015convolutional} and the influence relevance voter (IRV) method
of \citet{swamidass2009influence} (see Section~\ref{sec:other_comparisons}).
\tablename~\ref{table:other_comparisons} compares graph convolution models to
published results on these datasets under similar cross-validation conditions.
Graph convolution results were comparable to published NFP models, with
significant improvement on the photovoltaic efficiency task (note that the graph
convolution results are from multitask models trained on all three NFP datasets
while~\citet{duvenaud2015convolutional} report values for single-task models).
The 10-fold mean AUC and BEDROC scores on the HIV dataset were slightly lower than the published
IRV values. However, we held out 10\% of the data (one fold) in each
cross-validation round as a validation set for checkpoint selection, meaning
that the graph convolution models were trained with fewer examples than the
published IRV models.

\afterpage{
\begin{table*}[htbp]
    \caption{Comparison of graph convolution to neural fingerprint (NFP) and
    influence relevance voter (IRV) models. Section~\ref{sec:other_comparisons}
    provides details for datasets and experimental procedures. Note that the NFP
    comparisons were performed using multitask graph convolution models, and
    that graph convolution models for the HIV dataset were trained with fewer
    examples than IRV since one cross-validation fold was used as a held-out
    validation set.}
    \label{table:other_comparisons}
    \sisetup{separate-uncertainty=true}
    \centering
    \rowcolors{1}{}{lightgray}
    \begin{tabular}{ l l l S S }
    \toprule
    Model & Dataset & Metric & {Original} & {\makecell{Graph \\ Convolution}} \\
    \midrule
    \cellcolor{white} & Solubility (log M) & MSE & 0.52 \pm 0.07 & 0.46 \pm 0.08 \\
    \cellcolor{white} & Drug efficacy (nM EC$_{50}$) & MSE & 1.16 \pm 0.03 & 1.07 \pm 0.06 \\
    \multirow{-3}{*}{\cellcolor{white}NFP} & Photovoltaic efficiency (\%) & MSE & 1.43 \pm 0.09 & 1.10 \pm 0.06 \\
    \midrule
    \cellcolor{white} & \cellcolor{white} & AUC & 0.845 & 0.838 \pm 0.027 \\
    \multirow{-2}{*}{\cellcolor{white}IRV} &
        \multirow{-2}{*}{\cellcolor{white}HIV} &
        BEDROC ($\alpha=20$) & 0.630 & 0.613 \pm 0.048 \\
    \bottomrule
    \end{tabular}
\end{table*}
}

\subsection{Input featurization}
\label{sec:simple_features}

\begin{figure}[tb]
  % From code/plot_figures.py
  \centering
  \includegraphics[width=\linewidth]{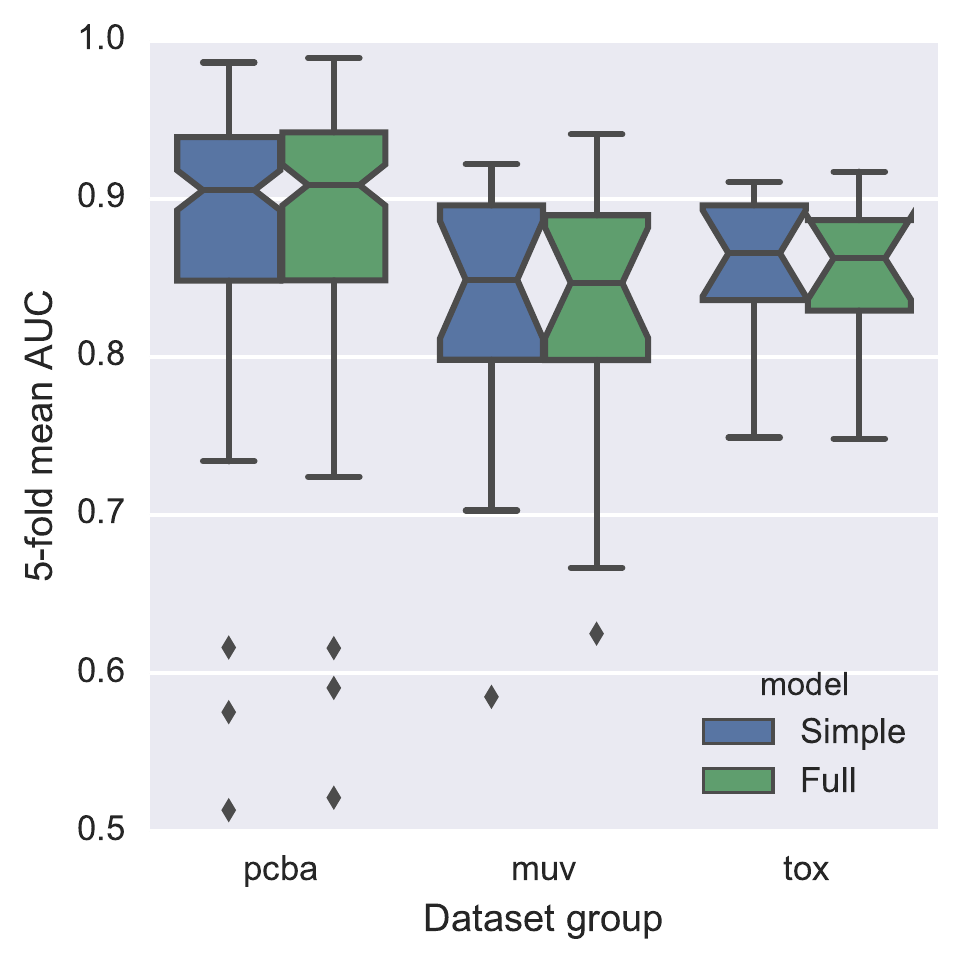}
  \caption{
    Comparison of models with ``simple'' and ``full'' input featurizations. The
    simple featurization only encodes atom type, bond type, and graph distance.
    The full featurization includes additional features such as aromaticity and
    hydrogen bonding propensity (see Section~\ref{sec:features} for more
    details). Confidence intervals for box plot medians were computed as
    ${\pm 1.57 \times \text{IQR} / \sqrt{N}}$ \citep{mcgill1978variations}.
  }
  \label{fig:simple_features}
\end{figure}

As a further proof of concept and to address the importance of the initial
featurization, we trained a model using a subset of features that match typical
2D structural diagrams seen in chemistry textbooks: only atom type, bond type,
and graph distance are provided to the network.
\figurename~\ref{fig:simple_features} compares a model trained with this
``simple'' input featurization to the ``full'' featurization containing all
features from \tablename~\ref{table:atom_features} and
\tablename~\ref{table:pair_features}. Both featurizations achieve similar median
5-fold mean AUC scores, suggesting that the additional features
in the ``full'' representation are either mostly ignored during training or can
be derived from a simpler representation of the molecular graph. Further work is
required to understand the importance of individual features, perhaps with
datasets that are sensitive to particular components of the input representation
(such as hydrogen bonding or formal charge).

\begin{figure*}[tb]
  % https://docs.google.com/drawings/d/1dpvd_lYvCw4IhKWSmz4bzKdjsZfKuQuENcmMMSCahQk/edit
  \centering
  \includegraphics[clip,trim=0 145 0 0,width=\linewidth]{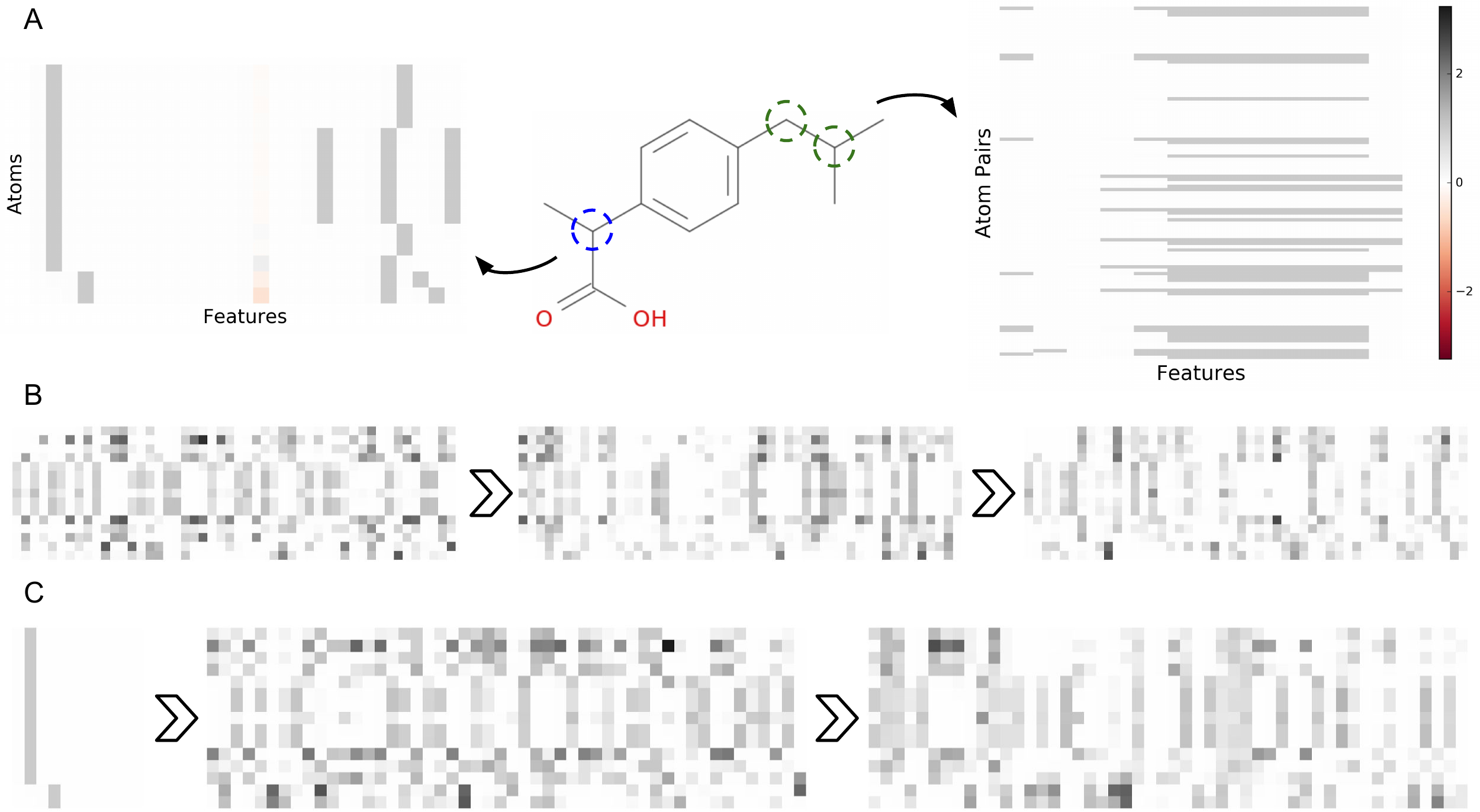}
  \caption{
    Graph convolution feature evolution. Atoms or pairs are
    displayed on the $y$-axis and the dimensions of the feature vectors are on
    the $x$-axis. (A) Conversion of the molecular graph
    for ibuprofen into atom and (unique) atom pair features. (B) Evolution of
    atom features after successive Weave modules in a graph convolution model
    with a W$_3$N$_2$ architecture and depth 50 convolutions in Weave modules.
    (C) Evolution of ``simple'' atom features (see
    Section~\ref{sec:simple_features}) starting from initial encoding and
    progressing through the Weave modules of a W$_2$N$_2$ architecture. The
    color bar applies to all panels.
  }
  \label{fig:feature_evolution}
\end{figure*}

\figurename~\ref{fig:feature_evolution} gives examples of how the initial atom
features for a single molecule (ibuprofen) evolve as they progress through graph
convolution Weave modules. The initial atom and pair feature encodings for the
``full'' featurization are depicted in Panel A. Comparing the initial atom
features to their source molecular graph, the aromatic carbons in the central
ring are clearly visible (and nearly identical in the featurization). The pair
features are more difficult to interpret visually, and mostly encode graph
distance. As the atom features are transformed by the Weave modules (Panel B),
they become more heterogeneous and reflective of their unique chemical
environments. ``Simple'' features behave similarly, beginning with rather
sterile initial values and quickly diverging as neighborhood information is
included by Weave module operations (Panel C). Comparison of the ``full'' and
``simple'' atom features after the second Weave module shows that both
featurizations lead to similarly diverse feature distributions.
\figurename~\ref{appendix:fig:feature_evolution_full} and
\figurename~\ref{appendix:fig:feature_evolution_simple} show similar behavior
for pair features.

\subsection{Hyperparameter sensitivity}
\label{sec:hyperparameter_sensitivity}

\subsubsection{Number of Weave modules}

\begin{figure}[tb]
  % From code/plot_figures.py
  \centering
  \includegraphics[width=\linewidth]{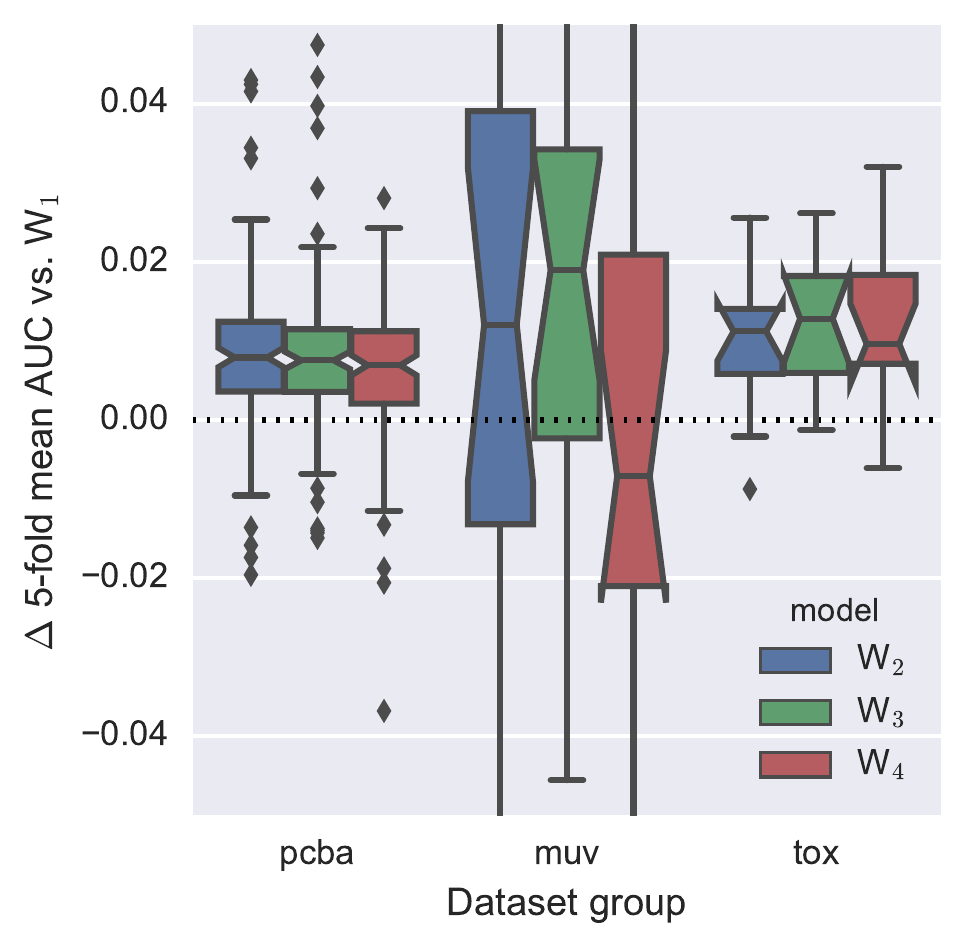}
  \caption{
    Comparison of models with different numbers of Weave modules with a model
    containing a single Weave module. All models used a maximum atom pair
    distance of two. The $y$-axis is cropped to emphasize differences near zero.
  }
  \label{fig:weave_delta}
\end{figure}

In relatively ``local'' models with limited atom pair distance, successive Weave
modules update atom features with information from progressively larger regions
of the molecule. This suggests that the number of Weave modules is a critical
hyperparameter to optimize, analogous to the number of hidden layers in
traditional neural networks. \figurename~\ref{fig:weave_delta} compares models
with 2--4 Weave modules to a model with a single Weave module. As expected,
models with a single Weave layer were outperformed by deeper architectures.
For the PCBA and Tox21 datasets, there was not much benefit to using more than
two Weave modules (\figurename~\ref{appendix:fig:weave}), but using
three Weave modules gave the best median AUC for the MUV datasets (in exchange
for significantly increased training time).

\subsubsection{Alternative feature reductions}

The reduction of atom features from the final Weave module to an
order-invariant, molecule-level representation is a major information bottleneck
in graph convolution models. In related work, a simple unweighted sum
\citep{duvenaud2015convolutional, merkwirth2005automatic,
lusci2013deep} or root-mean-square (RMS)~\citep{dieleman2015classifying}
reduction is used. Using a consistent base architecture with two Weave modules
and a maximum atom pair distance of 2, we
compared these traditional reduction strategies with our Gaussian histogram approach.

\begin{figure}[tb]
  % From code/plot_figures.py
  \centering
  \includegraphics[width=\linewidth]{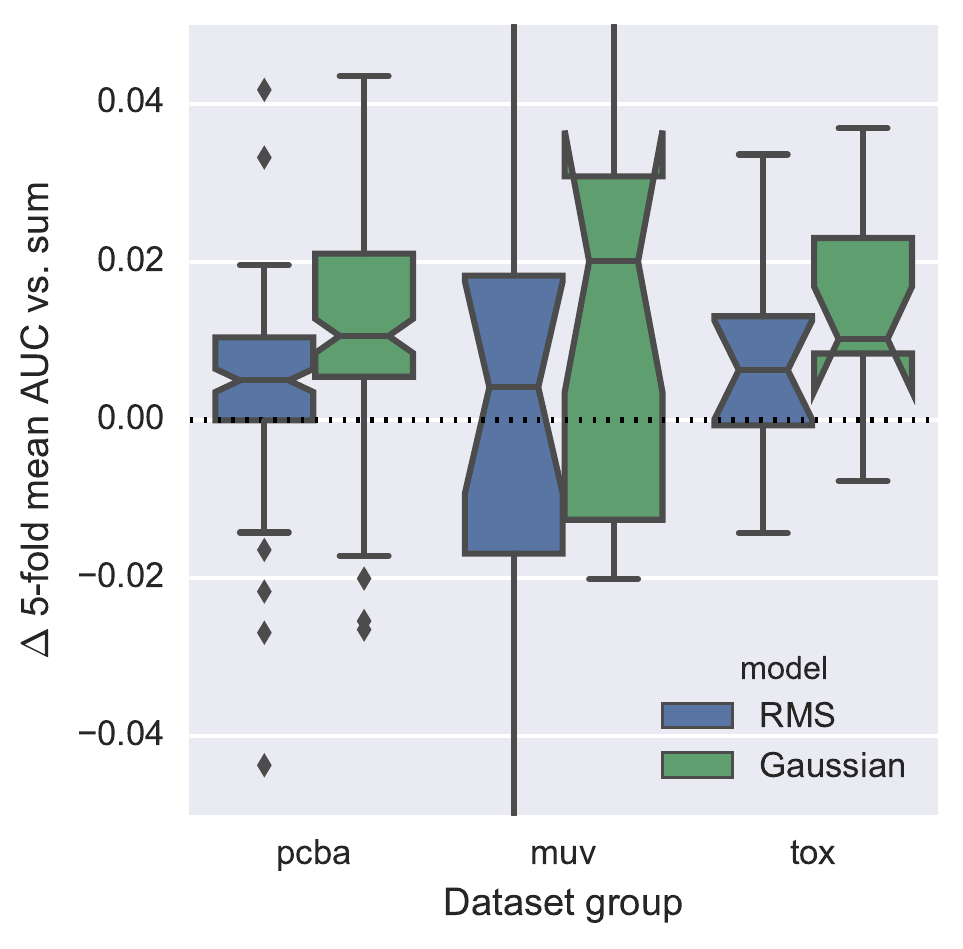}
  \caption{
    Comparison of root-mean-square (RMS) and Gaussian histogram reductions
    \emph{vs.} sum reduction. The $y$-axis reports difference in 5-fold mean AUC
    relative to sum reduction. All models used two Weave modules and a maximum
    atom pair distance of two. The $y$-axis is cropped to emphasize differences
    near zero.
  }
  \label{fig:reduction_delta}
\end{figure}

\figurename~\ref{fig:reduction_delta} shows that Gaussian histogram models had
consistently improved scores relative to sum reductions. RMS reductions were not
as robust as Gaussian histograms in terms of per-dataset differences relative to
sum reductions, although RMS and Gaussian histogram reductions had similar
distributions of absolute AUC values (\figurename~\ref{appendix:fig:reduction}).
Additionally, RMS reductions achieved a slightly higher median AUC than Gaussian
histogram reductions on the MUV datasets.

\subsubsection{Distance-dependent pair features}
\label{sec:neighbors}

In Weave modules, atoms are informed about their chemical environment by
mixing with pair features in the $P \rightarrow A$ operation. Recall that during
this operation, pair features are combined for pairs that contain a given
atom, yielding a new representation for that atom. A critical parameter for this
operation is the maximum distance (in bonds) allowed between the atoms of the
pairs that are combined. If only adjacent atoms are combined, the resulting
atom features will reflect the local chemical environment. As an alternative to
increasing the number of Weave modules, longer-range interactions can be
captured by increasing the maximum atom pair distance. However, our
implementation of the $P \rightarrow A$ operation uses a simple sum to combine
pair features, such that a large amount of information (possibly including every
pair of atoms in the molecule) is combined in a way that could prevent
useful information from being available in later stages of the network.

\begin{figure}[tb]
  % From code/plot_figures.py
  \centering
  \includegraphics[width=\linewidth]{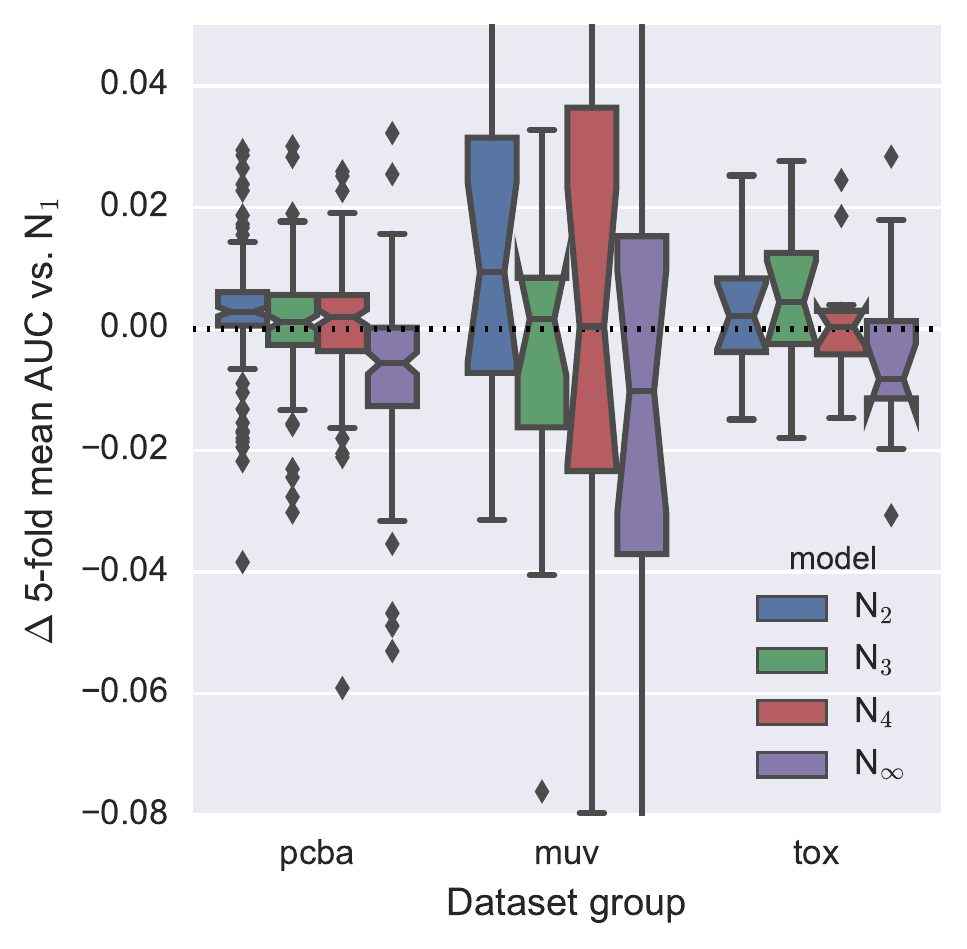}
  \caption{
    Comparison of models with different maximum atom pair distances to a model
    with a maximum pair distance of one (bonded atoms). All models have two
    Weave modules. The $y$-axis is cropped to emphasize differences near zero.
  }
  \label{fig:neighbors_delta}
\end{figure}

\figurename~\ref{fig:neighbors_delta} shows the performance of several models
with different maximum pair distances relative to a model that used only
adjacent atom pairs (N$_1$). For the PCBA datasets, a maximum distance of 2
(N$_2$) improves performance relative to the N$_1$ model, and N$_\infty$ (no
maximum distance) is clearly worse. However, the N$_1$ model achieves the best
median AUC score for the MUV and Tox21 datasets (\tablename~\ref{table:results}
and \figurename~\ref{appendix:fig:neighbors}). These results suggest that graph convolution models do not
effectively make use of the initial graph distance features to preserve or
emphasize distance-dependent information.

To further investigate the effect of distance information in Weave modules, we
experimented with models that use distance-specific weights for
operations involving pair features in order to maintain distance information
explicitly throughout the network. However, results for these models are preliminary and
were not included in this report.

%%%%%%%%%%%%%%%%%%%%%%%%%%%%%%%%%%%%%%%%%%%%%%%%%%%%%%%%%%%%%%%%%%%%%%%%%%%%%%%
\section{Discussion}\label{sec:discussion}

\emph{Graph convolutions} are a deep learning architecture for learning directly
from undirected graphs. In this work, we emphasize their application to small
molecules---undirected graphs of atoms connected by bonds---for virtual
screening. Starting from simple descriptions of atoms, bonds between atoms, and
pairwise relationships in a molecular graph, we have demonstrated performance
that is comparable to state of the art multitask neural networks trained on
traditional molecular fingerprint representations, as well as alternative
methods including ``neural fingerprints''~\citep{duvenaud2015convolutional} and
influence relevance voter~\citep{swamidass2009influence}.

Our experiments with the adjustable parameters in graph convolution models
indicate a relatively minor sensitivity to the number of Weave modules and the
maximum distance between atom pairs (at least for our datasets). These results suggest that a
model with two Weave modules, a maximum atom pair distance of 2, and
Gaussian histogram reductions is a good starting point for further optimization.
Remarkably, graph
convolution models perform well with a ``simple'' input featurization containing
only atom type, bond type, and graph distances---essentially the information
available from looking at \figurename~\ref{fig:ibuprofen}.

Flexibility is a highlight of the graph
convolution architecture: because we begin with a representation that encodes
the complete molecular graph, graph convolution models are free to use any of
the available information for the task at hand. In a sense, every possible
molecular ``fingerprint'' is available to the model.
Said another way, graph convolutions and other graph-based approaches
purposefully blur the distinction between molecular features and predictive
models. As has been pointed out elsewhere~\citep{duvenaud2015convolutional}, the
ability to use backpropagation to tune parameters at every stage of the network
provides greater representational power than traditional descriptors, which are
inflexible in the features they encode from the initial representation.
Accordingly, it is not appropriate to think of graph-based methods as
alternative descriptors; rather, they should be viewed as fully integrated
approaches to virtual screening (although future work could investigate the
utility of the learned molecule-level features for additional tasks or other
applications such as molecular similarity).

Looking forward, graph convolutions (and related graph-based methods; see
Section~\ref{sec:related_work}) present a ``new hill to climb'' in
computer-aided drug design and cheminformatics. Although our current graph
convolution models do not consistently outperform state-of-the-art
fingerprint-based models, we emphasize their flexibility and potential for
further optimization and development. In particular, we are aware of several
specific opportunities for improvement, including (1) additional optimization of
model hyperparameters such as Weave module convolution depths; (2) fine-tuning
of architectural decisions, such as the choice of reduction in the $P
\rightarrow A$ operation (currently a sum, but perhaps a Gaussian histogram or
distance-dependent function); and (3) improvements in memory usage and training
performance, such as not handling all pairs of atoms or implementing more
efficient versions of Weave module operations. With these and other
optimizations, we expect that graph convolutions could exceed the performance of
the best available fingerprint-based methods.

Finally, we note that much (or most) of the information required to represent
biological systems and the interactions responsible for small molecule activity
is not encapsulated in the molecular graph. Biology takes place in a
three-dimensional world, and is sensitive to shape, electrostatics, quantum
effects, and other properties that emerge from---but are not necessarily unique
to---the molecular graph (see, for example,~\citet{nicholls2010molecular}).
Additionally, most small molecules exhibit 3D conformational flexibility that
our graph representation does not even attempt to describe. The extension of
deep learning methods (including graph convolutions) to three-dimensional
biology is an active area of research (e.g.~\citet{wallach2015atomnet})
that requires special attention to the added complexities of multiple-instance
learning in a relatively small-data regime.

%%%%%%%%%%%%%%%%%%%%%%%%%%%%%%%%%%%%%%%%%%%%%%%%%%%%%%%%%%%%%%%%%%%%%%%%%%%%%%%

\section*{Acknowledgments}
%\begin{acknowledgements}

We thank Bharath Ramsundar, Brian Goldman, and Robert McGibbon for helpful
discussion. We also acknowledge Manjunath Kudlur, Derek Murray, and Rajat Monga
for assistance with TensorFlow. S.K. was supported by internships at Google Inc.
and Vertex Pharmaceuticals Inc. Additionally, we acknowledge use of the Stanford
BioX3 cluster supported by NIH S10 Shared Instrumentation Grant 1S10RR02664701.
S.K. and V.P. also acknowledge support from from NIH 5U19AI109662-02.

%\end{acknowledgements}

%auto-ignore

\section*{Version information}

Submitted to the Journal of Computer-Aided Molecular Design.
Comments on arXiv versions:

\textbf{v2:} Changed cross-validation scheme to use a held-out validation set
and made other changes in response to reviewer comments, such as including
comparisons to additional models and adding more background for the methods.

\textbf{v3:} Added ROC enrichment metrics and changed baseline model training
strategy to use sample weights. Added BEDROC comparison to IRV models. Corrected
an error in the logistic regression model training protocol and updated the
method used to calculate the number of unique molecules in our datasets. Some
AUC values changed slightly due to model retraining and/or reevaluation.

\putbib
\end{bibunit}

\appendix
\counterwithin{figure}{section}
\counterwithin{table}{section}
\onecolumn
\raggedbottom
\title{\textsc{Appendix}}
\author{}
\maketitle
\begin{bibunit}
%auto-ignore

\section{Appendix: Model comparison}

The following figures are box plot representations of the data summarized in
\tablename~\ref{table:results}, organized by dataset group. We provide (a) box
plots for absolute 5-fold mean AUC scores for each model and (b) difference box
plots showing differences in 5-fold mean AUC scores against the
pyramidal~(2000,~100) multitask neural network (PMTNN) baseline model. The
difference box plots are visual analogs of the sign test confidence intervals
reported in \tablename~\ref{table:results}. Note, however, that the confidence
intervals on box plot medians (calculated as $\pm 1.57 \times \text{IQR} /
\sqrt{N}$ \citep{mcgill1978variations}) do not necessarily correspond to the
sign test confidence intervals.

\begin{figure}[H]
  \centering
  \begin{subfigure}{0.49\linewidth}
    \includegraphics[width=\linewidth]{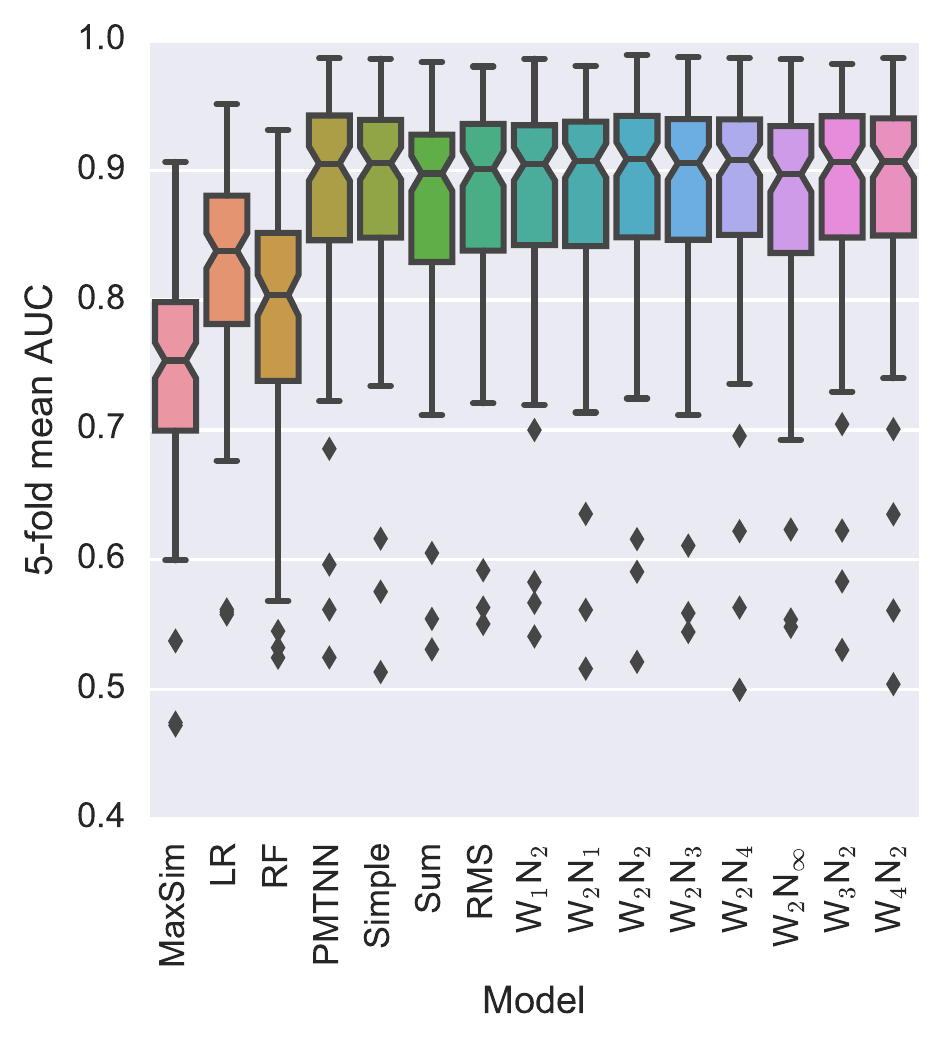}
    \caption{Full box plot.}
  \end{subfigure}
  \begin{subfigure}{0.49\linewidth}
    \includegraphics[width=\linewidth]{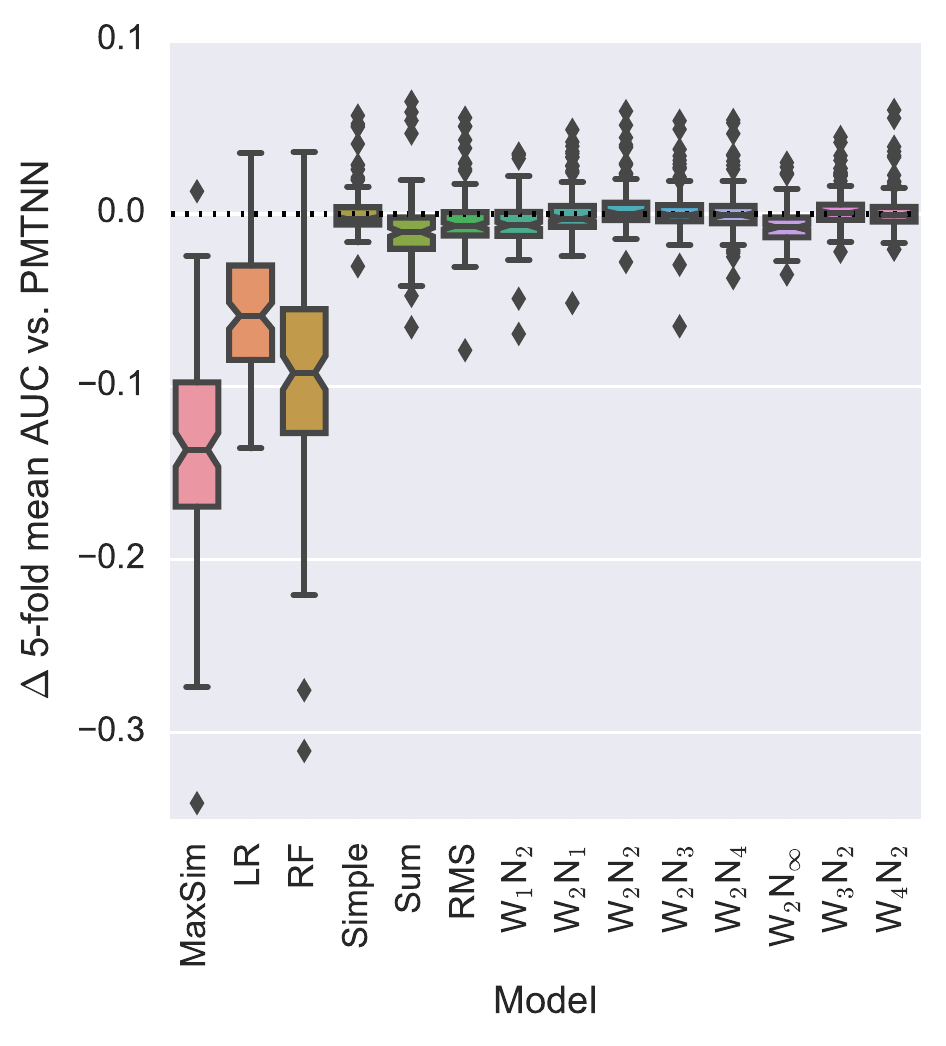}
    \caption{Difference box plot vs. PMTNN.}
  \end{subfigure}
  \caption{
    Model performance on PCBA datasets.
  }
  \label{appendix:fig:results_pcba}
\end{figure}

\begin{figure}[H]
  \centering
  \begin{subfigure}{0.49\linewidth}
    \includegraphics[width=\linewidth]{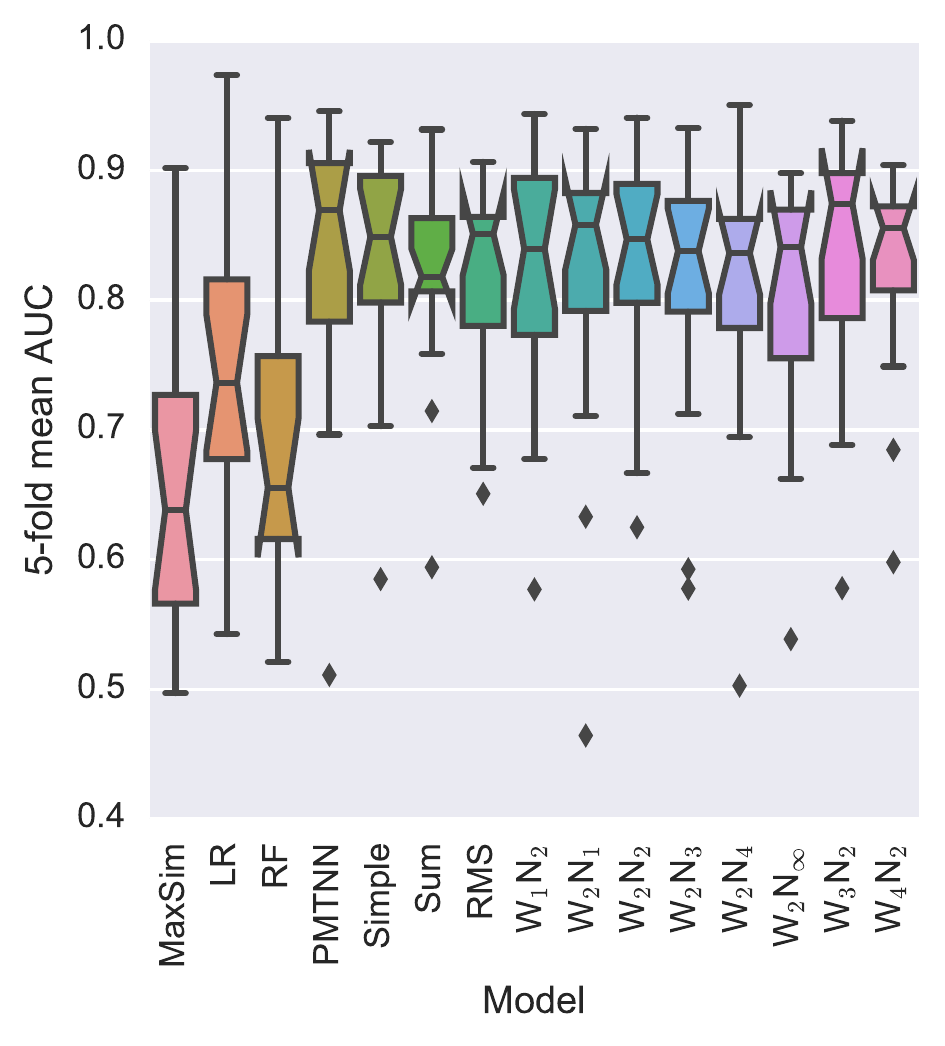}
    \caption{Full box plot.}
  \end{subfigure}
  \begin{subfigure}{0.49\linewidth}
    \includegraphics[width=\linewidth]{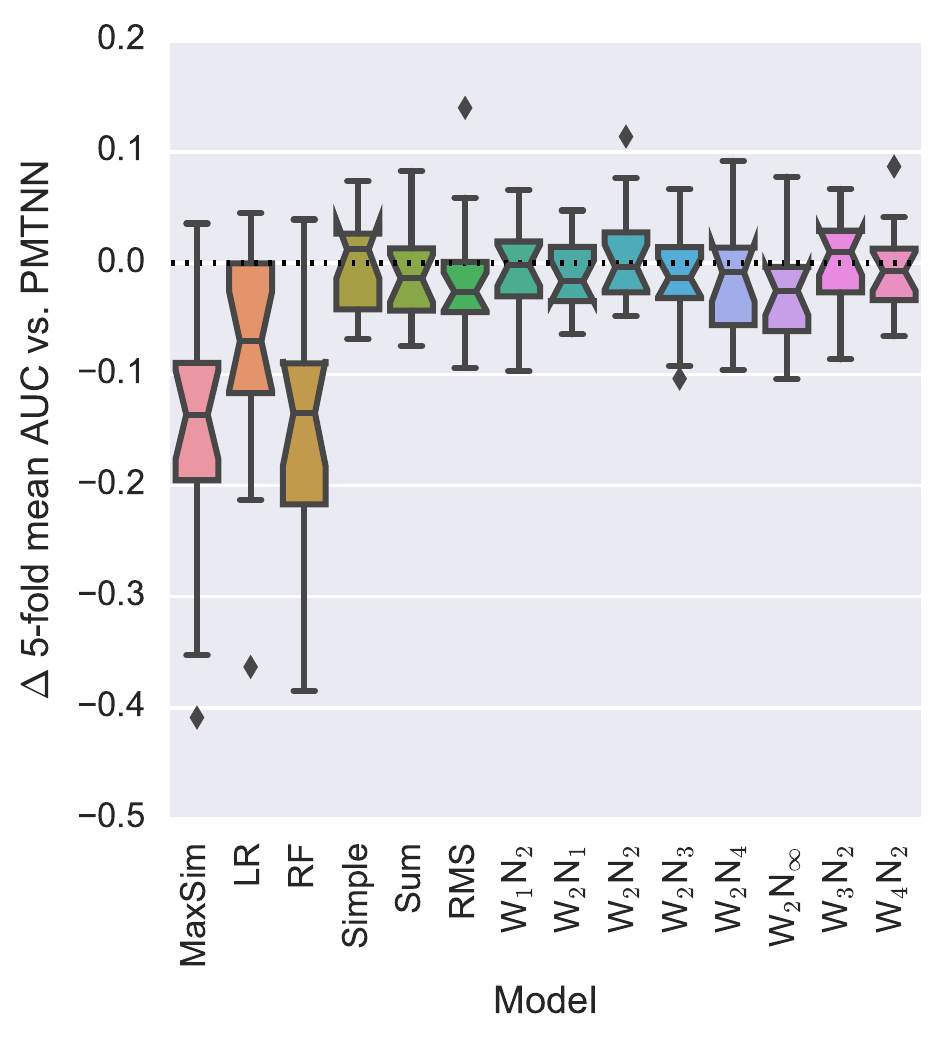}
    \caption{Difference box plot vs. PMTNN.}
  \end{subfigure}
  \caption{
    Model performance on MUV datasets.
  }
  \label{appendix:fig:results_muv}
\end{figure}

\begin{figure}[H]
  \centering
  \begin{subfigure}{0.49\linewidth}
    \includegraphics[width=\linewidth]{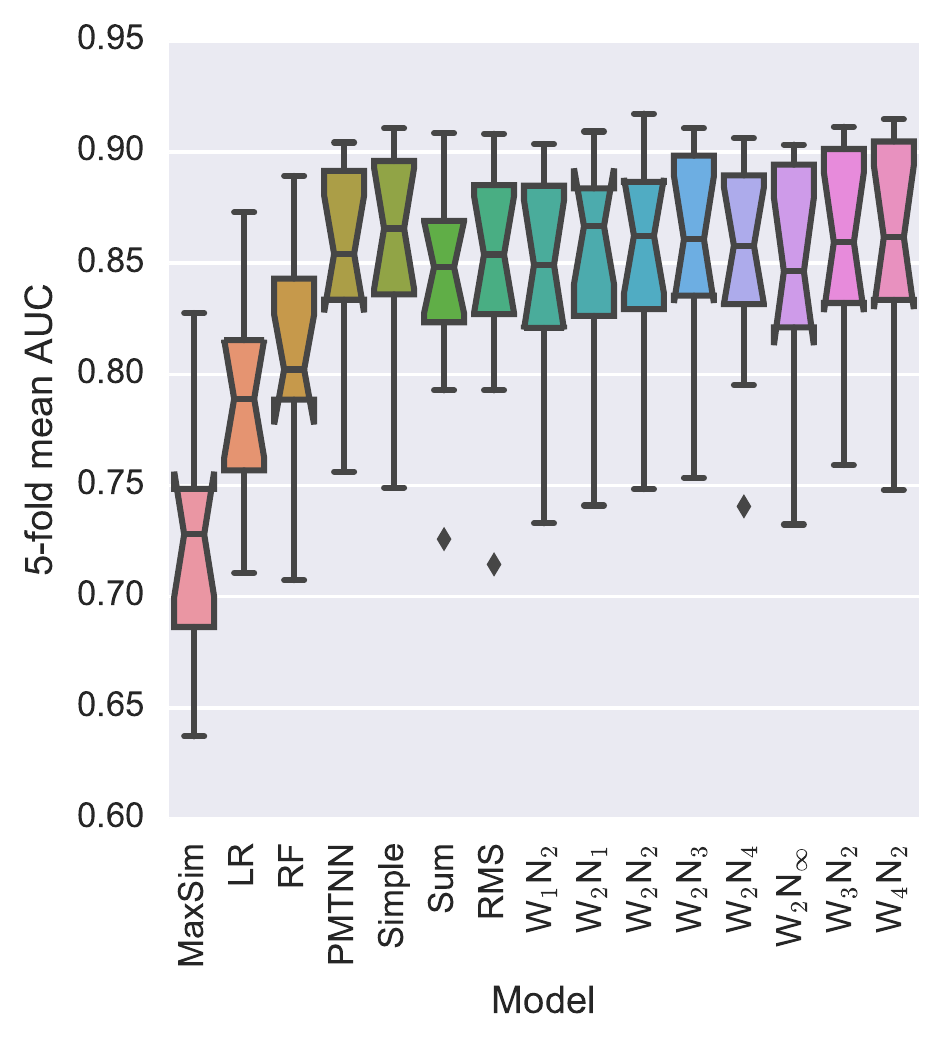}
    \caption{Full box plot.}
  \end{subfigure}
  \begin{subfigure}{0.49\linewidth}
    \includegraphics[width=\linewidth]{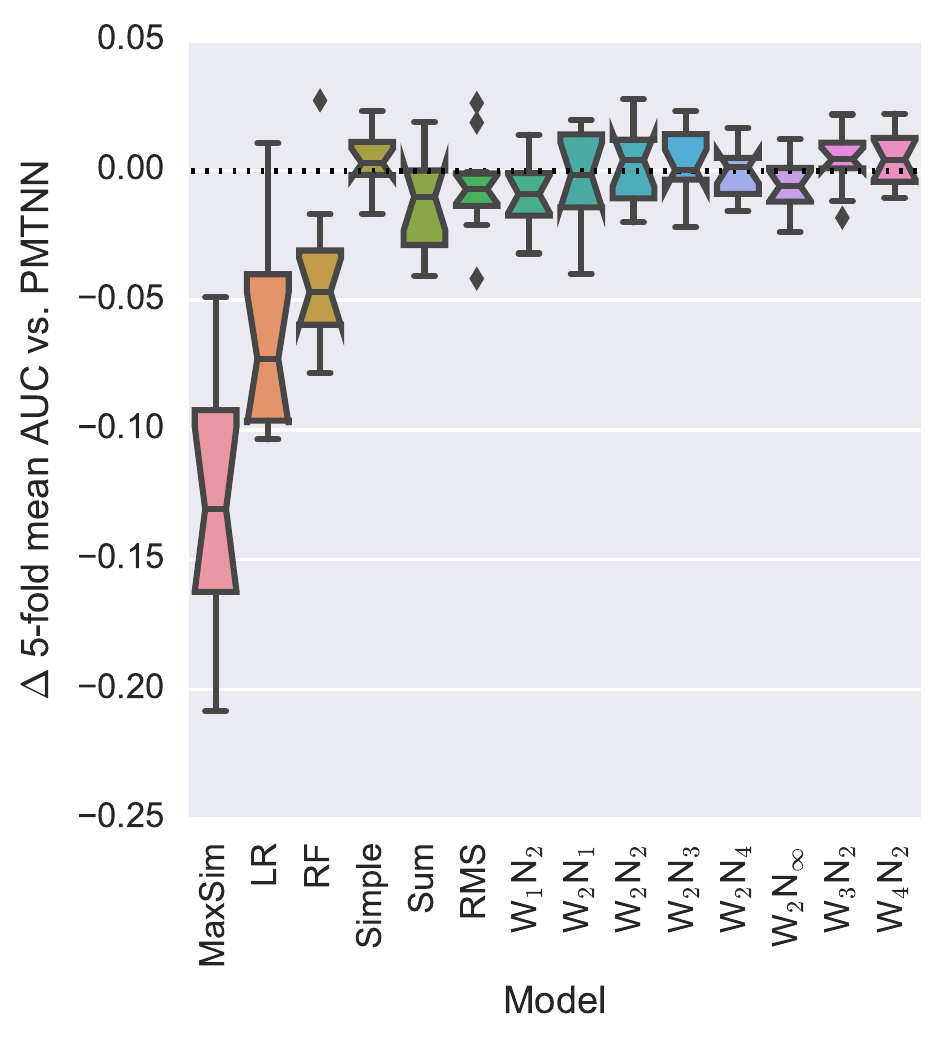}
    \caption{Difference box plot vs. PMTNN.}
  \end{subfigure}
  \caption{
    Model performance on Tox21 datasets.
  }
  \label{appendix:fig:results_tox}
\end{figure}

\pagebreak
\section{Appendix: ROC enrichment}
\label{sec:roc_enrichment}

The following tables report ROC enrichment~\citep{jain2008recommendations}
scores for baseline and graph convolution models. Each metric was optimized
separately using the held-out validation set for each model, such that ROC AUC
or ROC enrichment scores at different false positive rates (FPRs) are not
necessarily derived from predictions using the same set of model training
checkpoints.

\begin{landscape}
\vspace*{\fill}  % http://tex.stackexchange.com/questions/97050
\begin{table*}[htb]
    \caption{Median 5-fold mean ROC enrichment values for reported models at 1\%
    FPR ($E_{1\%}$). For each model, we report the median $\Delta E_{1\%}$ and
    the 95\% Wilson score interval for a sign test estimating the probability
    that a given model will outperform the PMTNN baseline (see
    Section~\ref{sec:training}). Bold values indicate sign test confidence
    intervals that do not include 0.5.}
    \label{table:roc_enrichment_1p}
    \centering
    \small
    \rowcolors{1}{}{lightgray}
    \begin{tabular}{ l S S c S S c S S c }
    \toprule
     & \multicolumn{3}{c}{PCBA $(n=128)$} &
       \multicolumn{3}{c}{MUV $(n=17)$} &
       \multicolumn{3}{c}{Tox21 $(n=12)$} \\
    \cmidrule(lr){2-4} \cmidrule(lr){5-7} \cmidrule(lr){8-10}
    Model & {\makecell{Median \\ $E_{1\%}$}} &
            {\makecell{Median \\ $\Delta E_{1\%}$}} &
            {\makecell{Sign Test \\ 95\% CI}} &
            {\makecell{Median \\ $E_{1\%}$}} &
            {\makecell{Median \\ $\Delta E_{1\%}$}} &
            {\makecell{Sign Test \\ 95\% CI}} &
            {\makecell{Median \\ $E_{1\%}$}} &
            {\makecell{Median \\ $\Delta E_{1\%}$}} &
            {\makecell{Sign Test \\ 95\% CI}} \\
    \midrule
    MaxSim & 24.1 & \bfseries \color{red} -16.2 & \bfseries \color{red} (\num{0.04}, \num{0.13}) & 13.3 & -3.3 & (\num{0.22}, \num{0.64}) & 12.8 & \bfseries \color{red} -13.0 & \bfseries \color{red} (\num{0.00}, \num{0.24}) \\
    LR & 20.2 & \bfseries \color{red} -18.8 & \bfseries \color{red} (\num{0.01}, \num{0.08}) & 16.7 & 0.0 & (\num{0.28}, \num{0.72}) & 17.8 & \bfseries \color{red} -5.1 & \bfseries \color{red} (\num{0.05}, \num{0.45}) \\
    RF & 34.5 & \bfseries \color{red} -6.9 & \bfseries \color{red} (\num{0.12}, \num{0.25}) & 23.3 & -3.3 & (\num{0.23}, \num{0.67}) & 26.4 & -0.2 & (\num{0.25}, \num{0.75}) \\
    PMTNN & 43.7 &  &  & 30.0 &  &  & 28.1 &  &  \\
    \midrule
    W$_2$N$_2$-simple & 42.3 & \bfseries \color{red} -1.6 & \bfseries \color{red} (\num{0.15}, \num{0.29}) & 30.0 & -3.3 & (\num{0.14}, \num{0.56}) & 24.7 & -1.1 & (\num{0.19}, \num{0.68}) \\
    W$_2$N$_2$-sum & 34.5 & \bfseries \color{red} -6.5 & \bfseries \color{red} (\num{0.05}, \num{0.15}) & 16.7 & \bfseries \color{red} -13.3 & \bfseries \color{red} (\num{0.03}, \num{0.36}) & 17.2 & \bfseries \color{red} -9.8 & \bfseries \color{red} (\num{0.01}, \num{0.35}) \\
    W$_2$N$_2$-RMS & 39.2 & \bfseries \color{red} -3.5 & \bfseries \color{red} (\num{0.04}, \num{0.14}) & 13.3 & \bfseries \color{red} -6.7 & \bfseries \color{red} (\num{0.01}, \num{0.30}) & 21.2 & \bfseries \color{red} -4.3 & \bfseries \color{red} (\num{0.05}, \num{0.45}) \\
    W$_1$N$_2$ & 38.3 & \bfseries \color{red} -3.6 & \bfseries \color{red} (\num{0.05}, \num{0.15}) & 20.0 & \bfseries \color{red} -3.3 & \bfseries \color{red} (\num{0.08}, \num{0.48}) & 22.6 & -4.7 & (\num{0.09}, \num{0.53}) \\
    W$_2$N$_1$ & 40.9 & \bfseries \color{red} -2.2 & \bfseries \color{red} (\num{0.17}, \num{0.31}) & 16.7 & -6.7 & (\num{0.14}, \num{0.56}) & 25.6 & -2.7 & (\num{0.09}, \num{0.53}) \\
    W$_2$N$_2$ & 42.2 & \bfseries \color{red} -0.8 & \bfseries \color{red} (\num{0.30}, \num{0.46}) & 26.7 & \bfseries \color{red} -3.3 & \bfseries \color{red} (\num{0.07}, \num{0.45}) & 26.2 & 1.6 & (\num{0.47}, \num{0.91}) \\
    W$_2$N$_3$ & 42.0 & \bfseries \color{red} -0.9 & \bfseries \color{red} (\num{0.18}, \num{0.33}) & 26.7 & \bfseries \color{red} -3.3 & \bfseries \color{red} (\num{0.10}, \num{0.49}) & 25.5 & 2.4 & (\num{0.39}, \num{0.86}) \\
    W$_2$N$_4$ & 42.0 & \bfseries \color{red} -0.7 & \bfseries \color{red} (\num{0.23}, \num{0.39}) & 23.3 & \bfseries \color{red} -6.7 & \bfseries \color{red} (\num{0.08}, \num{0.48}) & 23.5 & -0.4 & (\num{0.25}, \num{0.75}) \\
    W$_2$N$_\infty$ & 38.8 & \bfseries \color{red} -2.7 & \bfseries \color{red} (\num{0.06}, \num{0.17}) & 20.0 & -3.3 & (\num{0.14}, \num{0.56}) & 23.4 & -1.1 & (\num{0.09}, \num{0.53}) \\
    W$_3$N$_2$ & 42.1 & \bfseries \color{red} -1.0 & \bfseries \color{red} (\num{0.19}, \num{0.34}) & 26.7 & 0.0 & (\num{0.25}, \num{0.70}) & 24.8 & 0.5 & (\num{0.32}, \num{0.81}) \\
    W$_4$N$_2$ & 40.6 & \bfseries \color{red} -1.2 & \bfseries \color{red} (\num{0.22}, \num{0.38}) & 23.3 & \bfseries \color{red} -3.3 & \bfseries \color{red} (\num{0.08}, \num{0.48}) & 24.8 & -0.9 & (\num{0.09}, \num{0.53}) \\
    \bottomrule
    \end{tabular}
\end{table*}

\begin{table*}[htb]
    \caption{Median 5-fold mean ROC enrichment values for reported models at 5\%
    FPR ($E_{5\%}$). For each model, we report the median $\Delta E_{5\%}$ and
    the 95\% Wilson score interval for a sign test estimating the probability
    that a given model will outperform the PMTNN baseline (see
    Section~\ref{sec:training}). Bold values indicate sign test confidence
    intervals that do not include 0.5.}
    \label{table:roc_enrichment_5p}
    \centering
    \small
    \rowcolors{1}{}{lightgray}
    \begin{tabular}{ l S S c S S c S S c }
    \toprule
     & \multicolumn{3}{c}{PCBA $(n=128)$} &
       \multicolumn{3}{c}{MUV $(n=17)$} &
       \multicolumn{3}{c}{Tox21 $(n=12)$} \\
    \cmidrule(lr){2-4} \cmidrule(lr){5-7} \cmidrule(lr){8-10}
    Model & {\makecell{Median \\ $E_{5\%}$}} &
            {\makecell{Median \\ $\Delta E_{5\%}$}} &
            {\makecell{Sign Test \\ 95\% CI}} &
            {\makecell{Median \\ $E_{5\%}$}} &
            {\makecell{Median \\ $\Delta E_{5\%}$}} &
            {\makecell{Sign Test \\ 95\% CI}} &
            {\makecell{Median \\ $E_{5\%}$}} &
            {\makecell{Median \\ $\Delta E_{5\%}$}} &
            {\makecell{Sign Test \\ 95\% CI}} \\
    \midrule
    MaxSim & 8.5 & \bfseries \color{red} -4.4 & \bfseries \color{red} (\num{0.01}, \num{0.08}) & 6.0 & \bfseries \color{red} -3.3 & \bfseries \color{red} (\num{0.03}, \num{0.34}) & 6.7 & \bfseries \color{red} -3.9 & \bfseries \color{red} (\num{0.00}, \num{0.24}) \\
    LR & 8.8 & \bfseries \color{red} -3.6 & \bfseries \color{red} (\num{0.02}, \num{0.09}) & 6.0 & -2.0 & (\num{0.14}, \num{0.56}) & 8.3 & \bfseries \color{red} -1.9 & \bfseries \color{red} (\num{0.01}, \num{0.35}) \\
    RF & 10.2 & \bfseries \color{red} -2.5 & \bfseries \color{red} (\num{0.06}, \num{0.17}) & 6.0 & -2.0 & (\num{0.14}, \num{0.56}) & 9.6 & \bfseries \color{red} -1.0 & \bfseries \color{red} (\num{0.05}, \num{0.45}) \\
    PMTNN & 13.5 &  &  & 10.7 &  &  & 10.3 &  &  \\
    \midrule
    W$_2$N$_2$-simple & 13.4 & \bfseries \color{red} -0.3 & \bfseries \color{red} (\num{0.19}, \num{0.34}) & 10.0 & -1.3 & (\num{0.22}, \num{0.64}) & 10.1 & -0.2 & (\num{0.19}, \num{0.68}) \\
    W$_2$N$_2$-sum & 12.3 & \bfseries \color{red} -0.9 & \bfseries \color{red} (\num{0.12}, \num{0.25}) & 7.3 & \bfseries \color{red} -2.0 & \bfseries \color{red} (\num{0.04}, \num{0.38}) & 8.8 & \bfseries \color{red} -1.9 & \bfseries \color{red} (\num{0.01}, \num{0.35}) \\
    W$_2$N$_2$-RMS & 12.9 & \bfseries \color{red} -0.7 & \bfseries \color{red} (\num{0.12}, \num{0.25}) & 8.0 & \bfseries \color{red} -2.0 & \bfseries \color{red} (\num{0.06}, \num{0.41}) & 9.4 & \bfseries \color{red} -1.4 & \bfseries \color{red} (\num{0.01}, \num{0.35}) \\
    W$_1$N$_2$ & 13.0 & \bfseries \color{red} -0.5 & \bfseries \color{red} (\num{0.13}, \num{0.27}) & 9.3 & \bfseries \color{red} -2.0 & \bfseries \color{red} (\num{0.10}, \num{0.49}) & 9.9 & -0.8 & (\num{0.09}, \num{0.53}) \\
    W$_2$N$_1$ & 13.3 & \bfseries \color{red} -0.4 & \bfseries \color{red} (\num{0.20}, \num{0.35}) & 8.7 & \bfseries \color{red} -0.7 & \bfseries \color{red} (\num{0.01}, \num{0.33}) & 10.4 & -0.4 & (\num{0.14}, \num{0.61}) \\
    W$_2$N$_2$ & 13.6 & \bfseries \color{red} -0.1 & \bfseries \color{red} (\num{0.30}, \num{0.47}) & 10.0 & \bfseries \color{red} -1.3 & \bfseries \color{red} (\num{0.10}, \num{0.49}) & 10.4 & 0.0 & (\num{0.28}, \num{0.79}) \\
    W$_2$N$_3$ & 13.3 & \bfseries \color{red} -0.2 & \bfseries \color{red} (\num{0.24}, \num{0.40}) & 8.7 & -1.3 & (\num{0.12}, \num{0.55}) & 10.5 & -0.2 & (\num{0.19}, \num{0.68}) \\
    W$_2$N$_4$ & 13.3 & \bfseries \color{red} -0.2 & \bfseries \color{red} (\num{0.25}, \num{0.41}) & 8.7 & -1.3 & (\num{0.13}, \num{0.53}) & 10.2 & -0.2 & (\num{0.14}, \num{0.61}) \\
    W$_2$N$_\infty$ & 12.8 & \bfseries \color{red} -0.5 & \bfseries \color{red} (\num{0.06}, \num{0.16}) & 8.7 & \bfseries \color{red} -1.3 & \bfseries \color{red} (\num{0.03}, \num{0.34}) & 10.4 & -0.2 & (\num{0.15}, \num{0.65}) \\
    W$_3$N$_2$ & 13.6 & \bfseries \color{red} -0.1 & \bfseries \color{red} (\num{0.26}, \num{0.43}) & 9.3 & -0.0 & (\num{0.16}, \num{0.61}) & 10.4 & -0.2 & (\num{0.14}, \num{0.61}) \\
    W$_4$N$_2$ & 13.3 & \bfseries \color{red} -0.1 & \bfseries \color{red} (\num{0.29}, \num{0.46}) & 8.0 & -1.3 & (\num{0.14}, \num{0.56}) & 10.5 & -0.0 & (\num{0.25}, \num{0.75}) \\
    \bottomrule
    \end{tabular}
\end{table*}

\begin{table*}[htb]
    \caption{Median 5-fold mean ROC enrichment values for reported models at
    10\% FPR ($E_{10\%}$). For each model, we report the median $\Delta
    E_{10\%}$ and the 95\% Wilson score interval for a sign test estimating the
    probability that a given model will outperform the PMTNN baseline (see
    Section~\ref{sec:training}). Bold values indicate sign test confidence
    intervals that do not include 0.5.}
    \label{table:roc_enrichment_10p}
    \centering
    \small
    \rowcolors{1}{}{lightgray}
    \begin{tabular}{ l S S c S S c S S c }
    \toprule
     & \multicolumn{3}{c}{PCBA $(n=128)$} &
       \multicolumn{3}{c}{MUV $(n=17)$} &
       \multicolumn{3}{c}{Tox21 $(n=12)$} \\
    \cmidrule(lr){2-4} \cmidrule(lr){5-7} \cmidrule(lr){8-10}
    Model & {\makecell{Median \\ $E_{10\%}$}} &
            {\makecell{Median \\ $\Delta E_{10\%}$}} &
            {\makecell{Sign Test \\ 95\% CI}} &
            {\makecell{Median \\ $E_{10\%}$}} &
            {\makecell{Median \\ $\Delta E_{10\%}$}} &
            {\makecell{Sign Test \\ 95\% CI}} &
            {\makecell{Median \\ $E_{10\%}$}} &
            {\makecell{Median \\ $\Delta E_{10\%}$}} &
            {\makecell{Sign Test \\ 95\% CI}} \\
    \midrule
    MaxSim & 5.1 & \bfseries \color{red} -2.2 & \bfseries \color{red} (\num{0.00}, \num{0.06}) & 3.3 & \bfseries \color{red} -2.0 & \bfseries \color{red} (\num{0.04}, \num{0.38}) & 4.3 & \bfseries \color{red} -2.1 & \bfseries \color{red} (\num{0.00}, \num{0.24}) \\
    LR & 5.9 & \bfseries \color{red} -1.4 & \bfseries \color{red} (\num{0.01}, \num{0.08}) & 4.7 & -0.7 & (\num{0.26}, \num{0.69}) & 5.2 & \bfseries \color{red} -1.1 & \bfseries \color{red} (\num{0.00}, \num{0.24}) \\
    RF & 6.0 & \bfseries \color{red} -1.3 & \bfseries \color{red} (\num{0.04}, \num{0.14}) & 3.7 & -1.0 & (\num{0.13}, \num{0.53}) & 5.8 & \bfseries \color{red} -0.7 & \bfseries \color{red} (\num{0.05}, \num{0.45}) \\
    PMTNN & 7.8 &  &  & 6.3 &  &  & 6.4 &  &  \\
    \midrule
    W$_2$N$_2$-simple & 7.7 & \bfseries \color{red} -0.1 & \bfseries \color{red} (\num{0.26}, \num{0.42}) & 5.7 & -0.7 & (\num{0.15}, \num{0.58}) & 6.3 & 0.0 & (\num{0.25}, \num{0.75}) \\
    W$_2$N$_2$-sum & 7.2 & \bfseries \color{red} -0.4 & \bfseries \color{red} (\num{0.12}, \num{0.25}) & 5.3 & -0.7 & (\num{0.13}, \num{0.53}) & 5.9 & \bfseries \color{red} -0.6 & \bfseries \color{red} (\num{0.05}, \num{0.45}) \\
    W$_2$N$_2$-RMS & 7.5 & \bfseries \color{red} -0.2 & \bfseries \color{red} (\num{0.13}, \num{0.26}) & 5.3 & \bfseries \color{red} -1.0 & \bfseries \color{red} (\num{0.07}, \num{0.45}) & 5.9 & \bfseries \color{red} -0.4 & \bfseries \color{red} (\num{0.05}, \num{0.45}) \\
    W$_1$N$_2$ & 7.5 & \bfseries \color{red} -0.2 & \bfseries \color{red} (\num{0.12}, \num{0.25}) & 5.0 & \bfseries \color{red} -1.0 & \bfseries \color{red} (\num{0.10}, \num{0.49}) & 6.2 & \bfseries \color{red} -0.2 & \bfseries \color{red} (\num{0.05}, \num{0.45}) \\
    W$_2$N$_1$ & 7.6 & \bfseries \color{red} -0.1 & \bfseries \color{red} (\num{0.21}, \num{0.37}) & 6.0 & -0.7 & (\num{0.11}, \num{0.52}) & 6.3 & -0.1 & (\num{0.09}, \num{0.53}) \\
    W$_2$N$_2$ & 7.7 & \bfseries \color{red} -0.0 & \bfseries \color{red} (\num{0.28}, \num{0.44}) & 5.7 & -0.3 & (\num{0.18}, \num{0.61}) & 6.2 & 0.0 & (\num{0.25}, \num{0.75}) \\
    W$_2$N$_3$ & 7.7 & \bfseries \color{red} -0.0 & \bfseries \color{red} (\num{0.28}, \num{0.45}) & 5.7 & \bfseries \color{red} -0.7 & \bfseries \color{red} (\num{0.10}, \num{0.49}) & 6.3 & 0.1 & (\num{0.35}, \num{0.85}) \\
    W$_2$N$_4$ & 7.7 & \bfseries \color{red} -0.1 & \bfseries \color{red} (\num{0.25}, \num{0.41}) & 5.7 & -0.7 & (\num{0.13}, \num{0.53}) & 6.4 & 0.0 & (\num{0.25}, \num{0.75}) \\
    W$_2$N$_\infty$ & 7.4 & \bfseries \color{red} -0.3 & \bfseries \color{red} (\num{0.09}, \num{0.20}) & 5.0 & -1.0 & (\num{0.13}, \num{0.53}) & 6.3 & -0.1 & (\num{0.09}, \num{0.53}) \\
    W$_3$N$_2$ & 7.8 & -0.0 & (\num{0.34}, \num{0.51}) & 6.0 & -0.3 & (\num{0.17}, \num{0.59}) & 6.2 & -0.0 & (\num{0.25}, \num{0.75}) \\
    W$_4$N$_2$ & 7.7 & \bfseries \color{red} -0.0 & \bfseries \color{red} (\num{0.29}, \num{0.46}) & 5.7 & -0.7 & (\num{0.13}, \num{0.53}) & 6.3 & 0.1 & (\num{0.32}, \num{0.81}) \\
    \bottomrule
    \end{tabular}
\end{table*}

\begin{table*}[htb]
    \caption{Median 5-fold mean ROC enrichment values for reported models at
    20\% FPR ($E_{20\%}$). For each model, we report the median $\Delta
    E_{20\%}$ and the 95\% Wilson score interval for a sign test estimating the
    probability that a given model will outperform the PMTNN baseline (see
    Section~\ref{sec:training}). Bold values indicate sign test confidence
    intervals that do not include 0.5.}
    \label{table:roc_enrichment_20p}
    \centering
    \small
    \rowcolors{1}{}{lightgray}
    \begin{tabular}{ l S S c S S c S S c }
    \toprule
     & \multicolumn{3}{c}{PCBA $(n=128)$} &
       \multicolumn{3}{c}{MUV $(n=17)$} &
       \multicolumn{3}{c}{Tox21 $(n=12)$} \\
    \cmidrule(lr){2-4} \cmidrule(lr){5-7} \cmidrule(lr){8-10}
    Model & {\makecell{Median \\ $E_{20\%}$}} &
            {\makecell{Median \\ $\Delta E_{20\%}$}} &
            {\makecell{Sign Test \\ 95\% CI}} &
            {\makecell{Median \\ $E_{20\%}$}} &
            {\makecell{Median \\ $\Delta E_{20\%}$}} &
            {\makecell{Sign Test \\ 95\% CI}} &
            {\makecell{Median \\ $E_{20\%}$}} &
            {\makecell{Median \\ $\Delta E_{20\%}$}} &
            {\makecell{Sign Test \\ 95\% CI}} \\
    \midrule
    MaxSim & 3.0 & \bfseries \color{red} -1.1 & \bfseries \color{red} (\num{0.00}, \num{0.03}) & 2.2 & \bfseries \color{red} -1.0 & \bfseries \color{red} (\num{0.03}, \num{0.34}) & 2.8 & \bfseries \color{red} -1.1 & \bfseries \color{red} (\num{0.00}, \num{0.24}) \\
    LR & 3.6 & \bfseries \color{red} -0.5 & \bfseries \color{red} (\num{0.03}, \num{0.11}) & 3.0 & -0.5 & (\num{0.18}, \num{0.61}) & 3.2 & \bfseries \color{red} -0.5 & \bfseries \color{red} (\num{0.01}, \num{0.35}) \\
    RF & 3.4 & \bfseries \color{red} -0.7 & \bfseries \color{red} (\num{0.03}, \num{0.11}) & 2.5 & \bfseries \color{red} -0.7 & \bfseries \color{red} (\num{0.03}, \num{0.36}) & 3.4 & \bfseries \color{red} -0.4 & \bfseries \color{red} (\num{0.01}, \num{0.35}) \\
    PMTNN & 4.2 &  &  & 3.8 &  &  & 3.7 &  &  \\
    \midrule
    W$_2$N$_2$-simple & 4.3 & \bfseries \color{red} -0.0 & \bfseries \color{red} (\num{0.30}, \num{0.46}) & 3.3 & \bfseries \color{red} -0.3 & \bfseries \color{red} (\num{0.10}, \num{0.49}) & 3.8 & 0.0 & (\num{0.32}, \num{0.81}) \\
    W$_2$N$_2$-sum & 4.2 & \bfseries \color{red} -0.1 & \bfseries \color{red} (\num{0.17}, \num{0.31}) & 3.3 & \bfseries \color{red} -0.3 & \bfseries \color{red} (\num{0.07}, \num{0.43}) & 3.7 & -0.1 & (\num{0.09}, \num{0.53}) \\
    W$_2$N$_2$-RMS & 4.2 & \bfseries \color{red} -0.1 & \bfseries \color{red} (\num{0.19}, \num{0.34}) & 3.5 & -0.2 & (\num{0.11}, \num{0.52}) & 3.8 & -0.1 & (\num{0.09}, \num{0.53}) \\
    W$_1$N$_2$ & 4.2 & \bfseries \color{red} -0.1 & \bfseries \color{red} (\num{0.19}, \num{0.34}) & 3.7 & -0.3 & (\num{0.14}, \num{0.56}) & 3.7 & -0.0 & (\num{0.14}, \num{0.61}) \\
    W$_2$N$_1$ & 4.3 & \bfseries \color{red} -0.0 & \bfseries \color{red} (\num{0.32}, \num{0.49}) & 3.5 & -0.2 & (\num{0.23}, \num{0.67}) & 3.9 & 0.0 & (\num{0.25}, \num{0.75}) \\
    W$_2$N$_2$ & 4.3 & -0.0 & (\num{0.38}, \num{0.55}) & 3.5 & -0.3 & (\num{0.17}, \num{0.59}) & 3.9 & 0.1 & (\num{0.35}, \num{0.85}) \\
    W$_2$N$_3$ & 4.3 & -0.0 & (\num{0.35}, \num{0.52}) & 3.3 & -0.3 & (\num{0.26}, \num{0.69}) & 3.8 & 0.0 & (\num{0.32}, \num{0.81}) \\
    W$_2$N$_4$ & 4.3 & \bfseries \color{red} -0.0 & \bfseries \color{red} (\num{0.28}, \num{0.45}) & 3.3 & \bfseries \color{red} -0.3 & \bfseries \color{red} (\num{0.10}, \num{0.47}) & 3.8 & -0.0 & (\num{0.25}, \num{0.75}) \\
    W$_2$N$_\infty$ & 4.2 & \bfseries \color{red} -0.1 & \bfseries \color{red} (\num{0.12}, \num{0.25}) & 3.3 & \bfseries \color{red} -0.3 & \bfseries \color{red} (\num{0.07}, \num{0.43}) & 3.8 & -0.0 & (\num{0.19}, \num{0.68}) \\
    W$_3$N$_2$ & 4.3 & -0.0 & (\num{0.37}, \num{0.54}) & 3.5 & -0.2 & (\num{0.23}, \num{0.67}) & 3.8 & 0.1 & (\num{0.32}, \num{0.81}) \\
    W$_4$N$_2$ & 4.3 & -0.0 & (\num{0.34}, \num{0.51}) & 3.7 & -0.2 & (\num{0.16}, \num{0.61}) & 3.8 & 0.1 & (\num{0.47}, \num{0.91}) \\
    \bottomrule
    \end{tabular}
\end{table*}
\vspace*{\fill}
\end{landscape}

\pagebreak
\section{Appendix: Input featurization}

For each of the experiments described in Section~\ref{sec:simple_features},
we provide figures showing (a) box plots for absolute 5-fold mean AUC scores for
each model and (b) difference box plots showing differences in 5-fold mean AUC
scores against a baseline model (without any $y$-axis cropping).

\begin{figure}[H]
  % From code/plot_figures.py
  \centering
  \begin{subfigure}{0.49\linewidth}
    \includegraphics[width=\linewidth]{simple-box.pdf}
    \caption{Full box plot.}
  \end{subfigure}
  \begin{subfigure}{0.49\linewidth}
    \includegraphics[width=\linewidth]{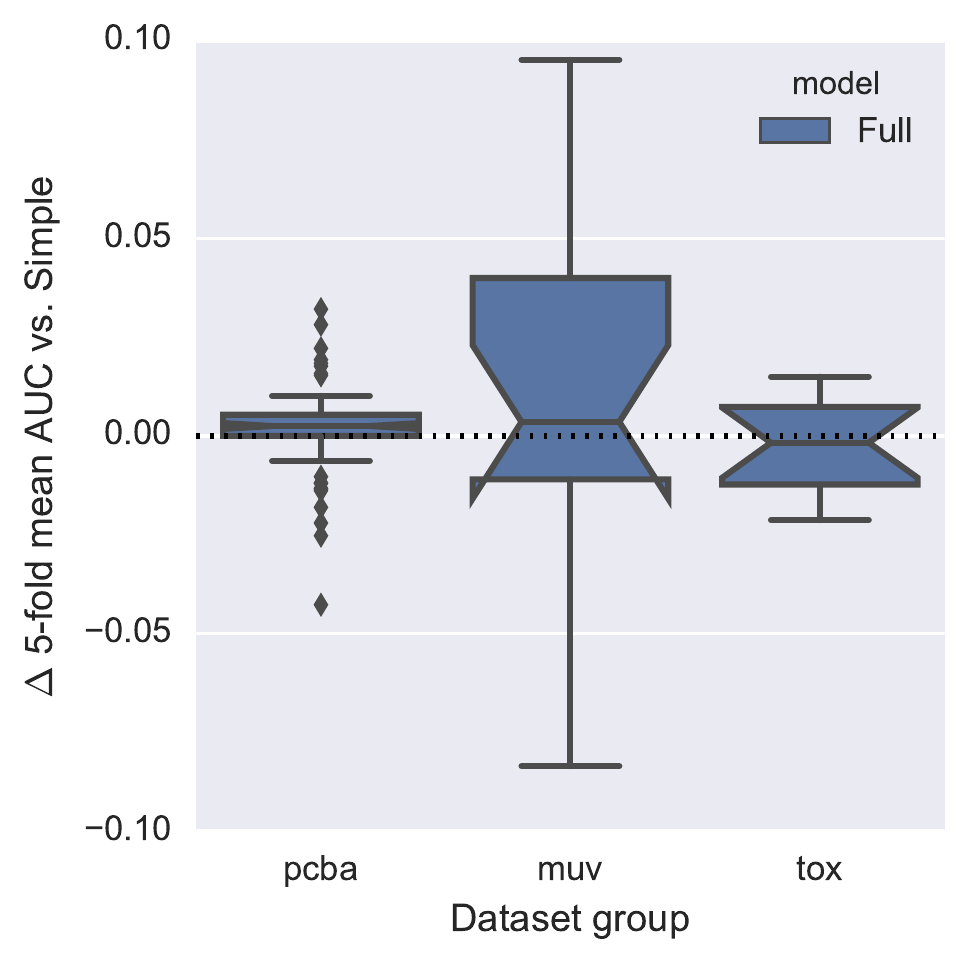}
    \caption{Difference box plot vs. ``simple'' featurization.}
  \end{subfigure}
  \caption{
    Comparison of models with ``simple'' and ``full'' input featurizations.
  }
  \label{appendix:fig:simple}
\end{figure}

\pagebreak
\section{Appendix: Hyperparameter sensitivity}

For each of the experiments described in
Section~\ref{sec:hyperparameter_sensitivity}, we provide figures showing
(a) box plots for absolute 5-fold mean AUC scores for each model and (b)
difference box plots showing differences in 5-fold mean AUC scores against a
baseline model (without any $y$-axis cropping).

\subsection{Number of Weave modules}

\begin{figure}[H]
  % From code/plot_figures.py
  \centering
  \begin{subfigure}{0.49\linewidth}
    \includegraphics[width=\linewidth]{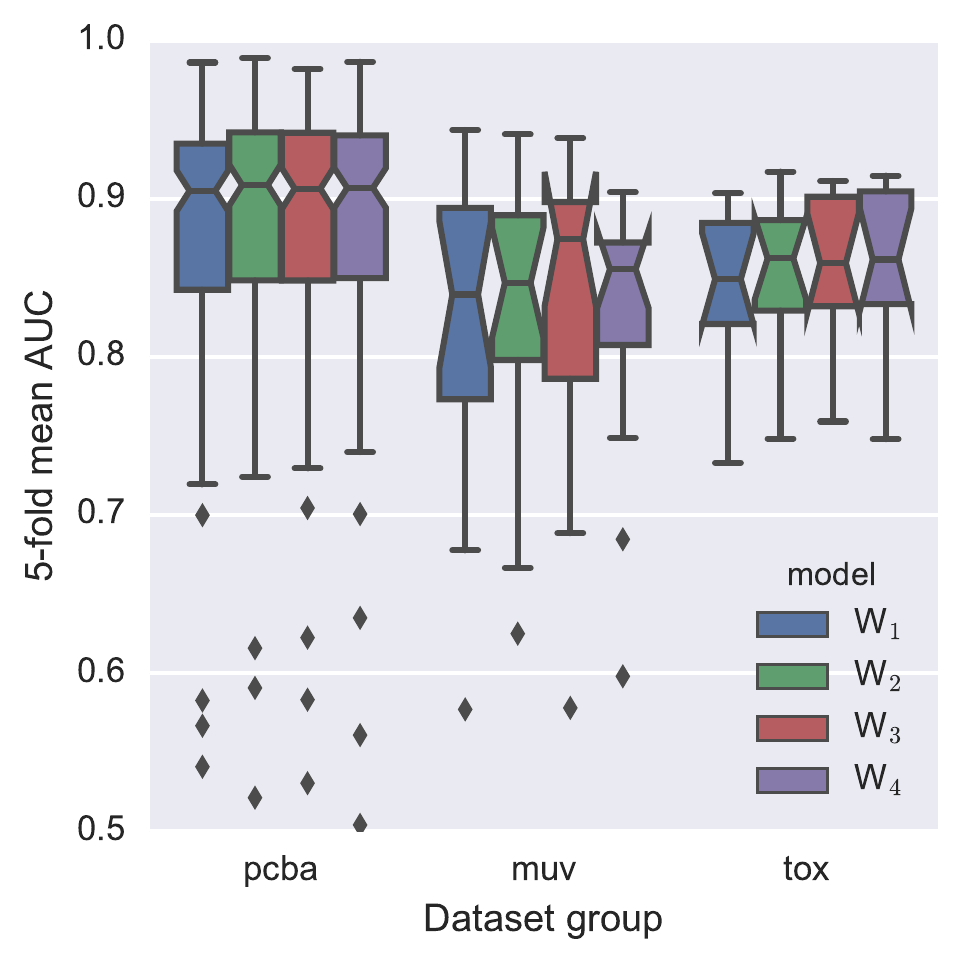}
    \caption{Full box plot.}
  \end{subfigure}
  \begin{subfigure}{0.49\linewidth}
    \includegraphics[width=\linewidth]{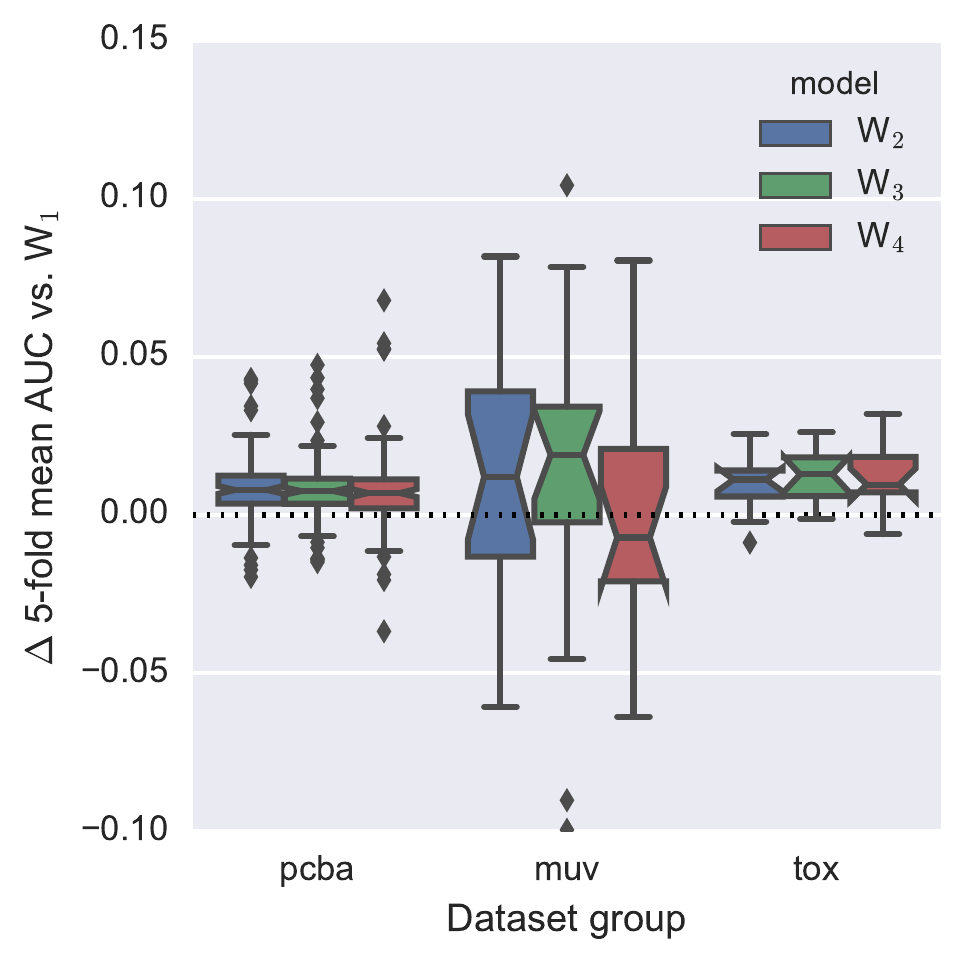}
    \caption{Difference box plot vs. W$_1$ model.}
  \end{subfigure}
  \caption{
    Comparison of models with different numbers of Weave modules.
  }
  \label{appendix:fig:weave}
\end{figure}

\subsection{Alternative feature reductions}

\begin{figure}[H]
  % From code/plot_figures.py
  \centering
  \begin{subfigure}{0.49\linewidth}
    \includegraphics[width=\linewidth]{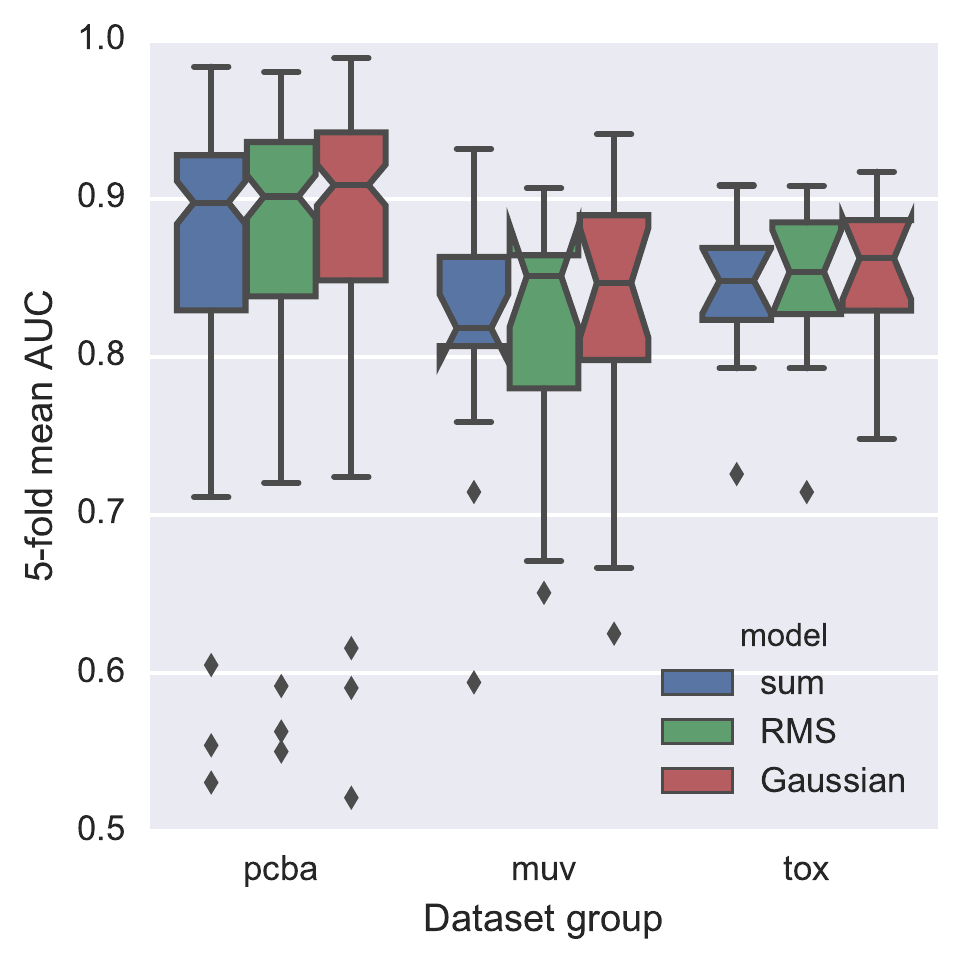}
    \caption{Full box plot.}
  \end{subfigure}
  \begin{subfigure}{0.49\linewidth}
    \includegraphics[width=\linewidth]{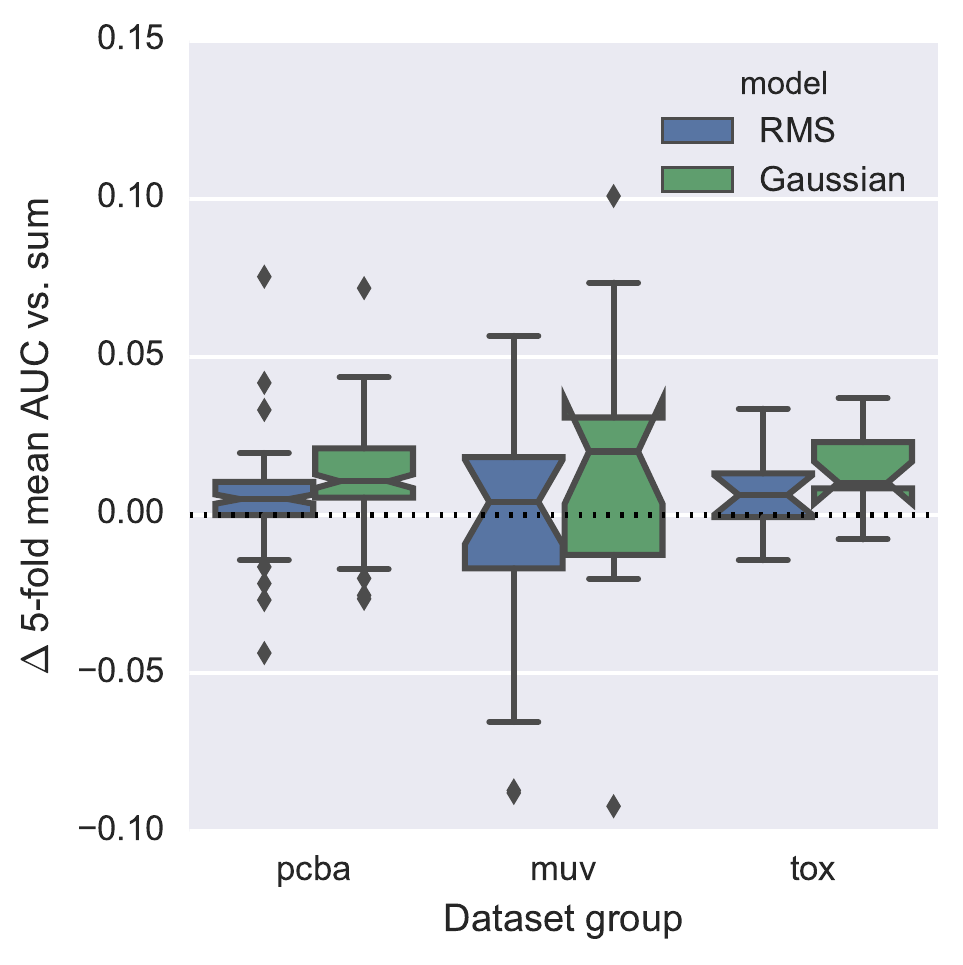}
    \caption{Difference box plot vs. sum reduction.}
  \end{subfigure}
  \caption{
    Comparison of models with different feature reduction methods.
  }
  \label{appendix:fig:reduction}
\end{figure}

\subsection{Distance-dependent pair features}

\begin{figure}[H]
  % From code/plot_figures.py
  \centering
  \begin{subfigure}{0.49\linewidth}
    \includegraphics[width=\linewidth]{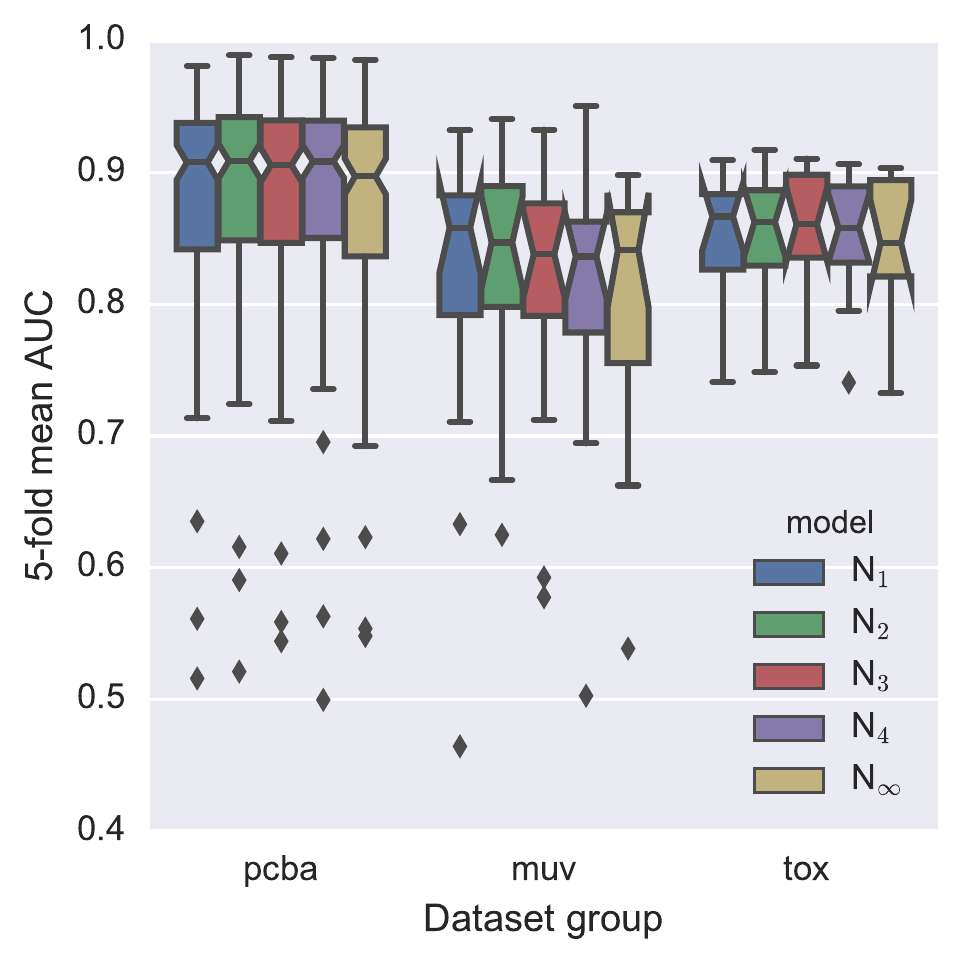}
    \caption{Full box plot.}
  \end{subfigure}
  \begin{subfigure}{0.49\linewidth}
    \includegraphics[width=\linewidth]{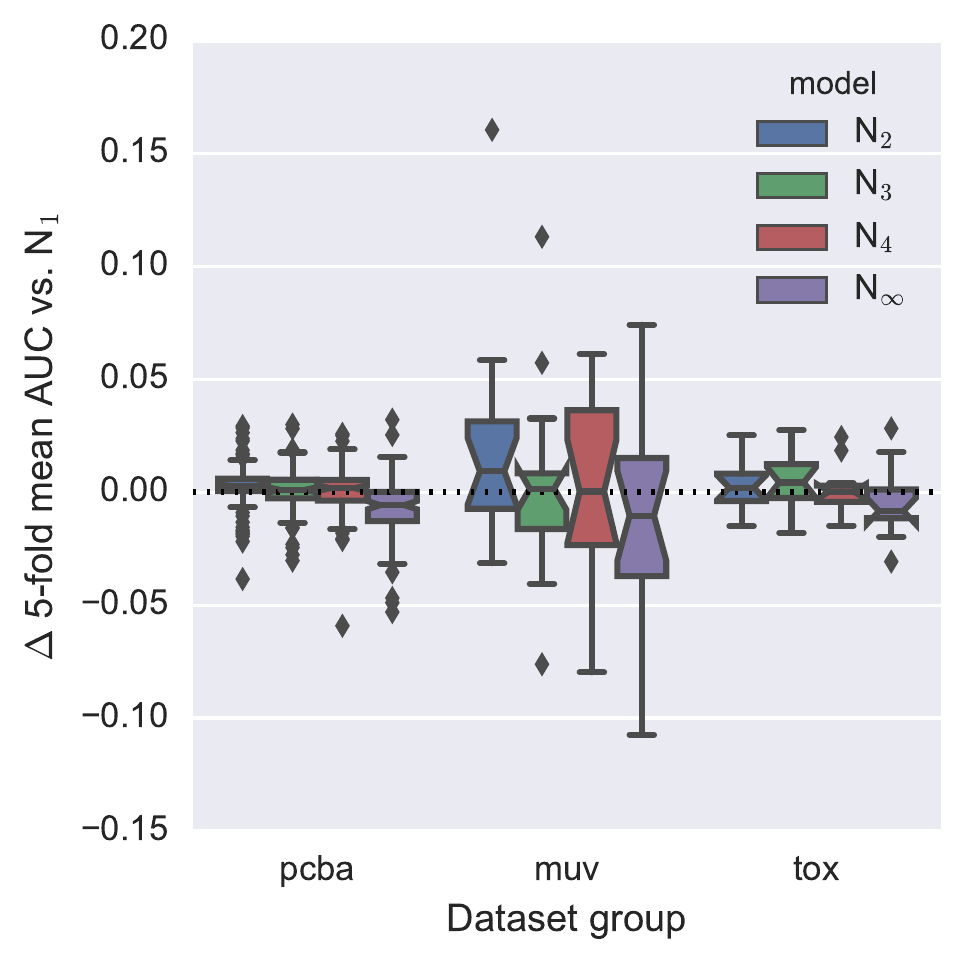}
    \caption{Difference box plot vs. N$_1$ model.}
  \end{subfigure}
  \caption{
    Comparison of models with different maximum atom pair distances.
  }
  \label{appendix:fig:neighbors}
\end{figure}

\pagebreak
\section{Appendix: Atom pair feature evolution}

\figurename~\ref{fig:feature_evolution} showed the evolution of atom
features at different stages of a graph convolution model (after subsequent
Weave modules). The following figures show the evolution of atom pair features
from the same models, using both the ``full'' and ``simple'' input
featurization. As in \figurename~\ref{fig:feature_evolution}, the initial
pair features describe ibuprofen. Most of the initial featurization describes
the graph distance between the atoms in the pair (see
\tablename~\ref{table:pair_features}). There are many blank rows since
pairs separated by more than the maximum atom pair distance are masked. Note
that only unique pairs are represented (i.e. $(a, b)$ but not $(b, a)$).
As the pair features move through the graph convolution network, it can be seen
that similar initial featurizations diverge as a consequence of Weave module
operations.

\begin{figure}[H]
  % From
  % https://docs.google.com/drawings/d/1KpUWwN3Igca4TEvY1uSoxD-cqJL1BXhaNOpm5wJ51V0/edit
  \centering
  \includegraphics[clip,trim=0 160 0 0,width=\linewidth]{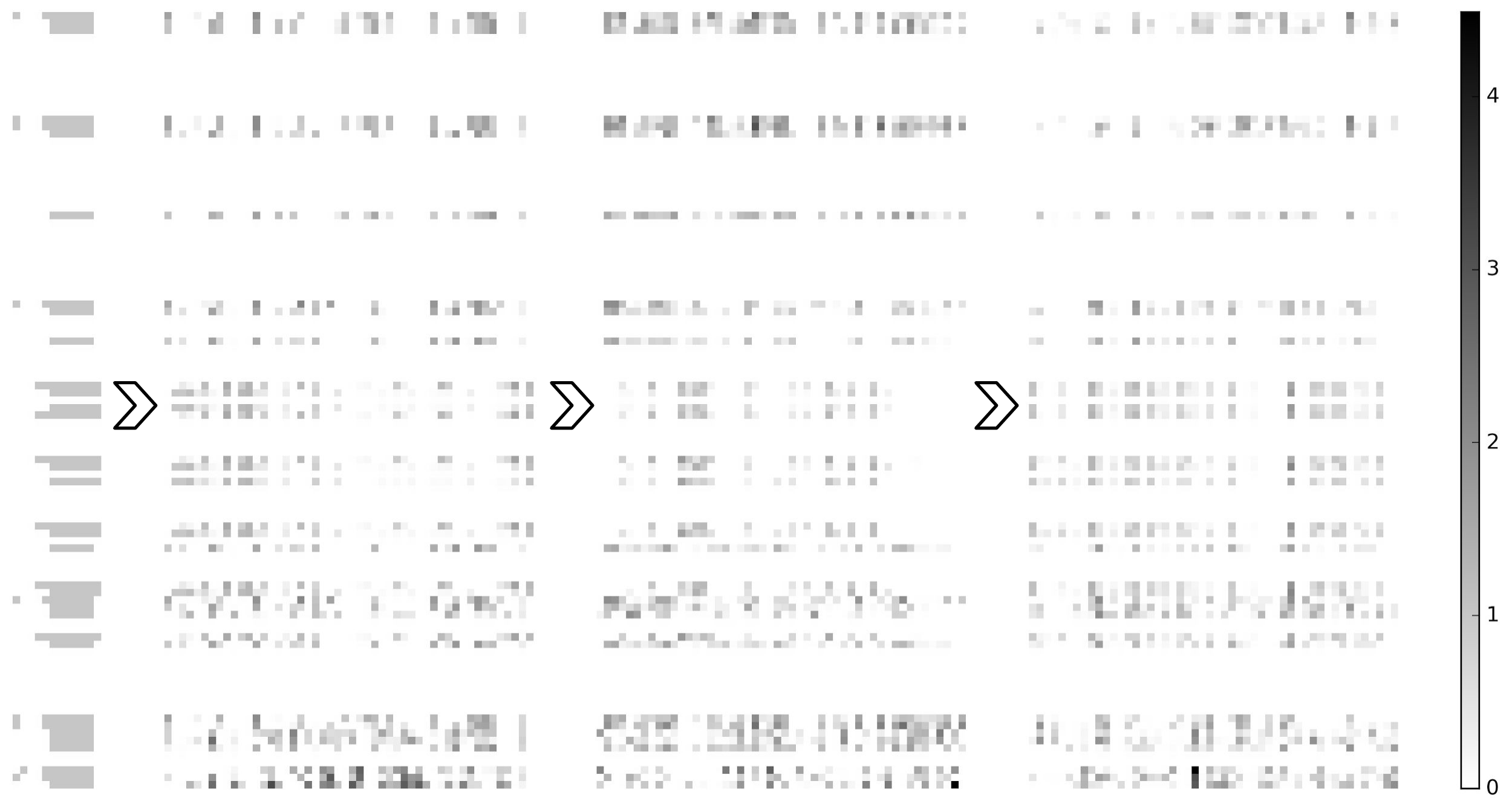}
  \caption{
    Graph convolution atom pair feature evolution using the ``full''
    featurization in a W$_3$N$_2$ architecture. Unique atom pairs are on the
    $y$-axis (one atom pair per row). Initial pair features are shown on the
    left, with whitespace separating subsequent Weave module outputs.
  }
  \label{appendix:fig:feature_evolution_full}
\end{figure}

\begin{figure}[H]
  % From
  % https://docs.google.com/drawings/d/1_JELnA79ieoB31FiZIDzzp62-S3MTYYqfmU3VIjMH-I/edit
  \centering
  \includegraphics[clip,trim=0 0 50 0,width=\linewidth]{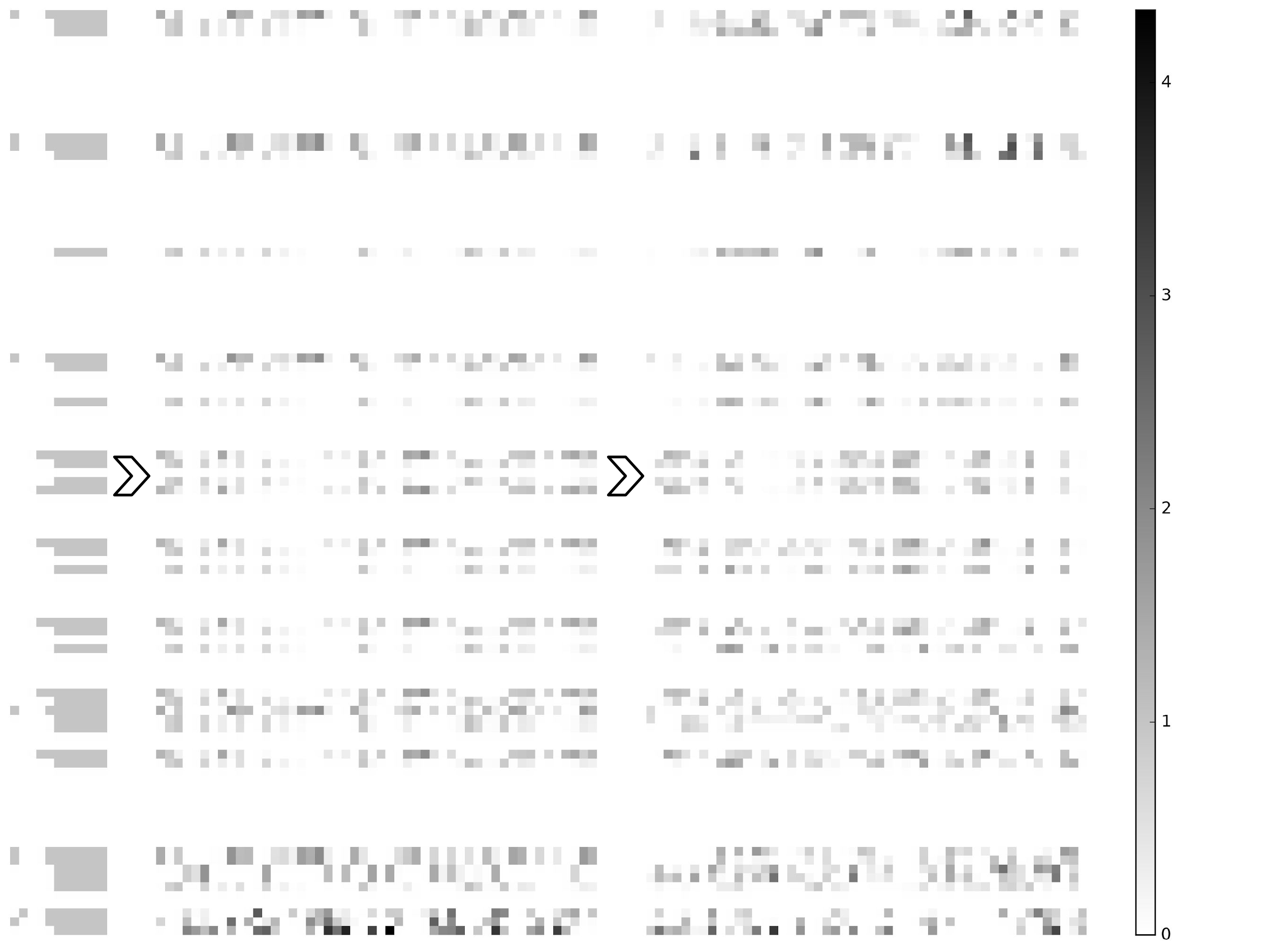}
  \caption{
    Graph convolution atom pair feature evolution using the ``simple''
    featurization in a W$_2$N$_2$ architecture. Unique atom pairs are on the
    $y$-axis (one atom pair per row). Initial pair features are shown on the
    left, with whitespace separating subsequent Weave module outputs.
  }
  \label{appendix:fig:feature_evolution_simple}
\end{figure}

\pagebreak
\section{Appendix: Gaussian histogram membership functions}

\begin{table*}[htbp]
  \caption{Gaussian membership functions.}
  \label{appendix:table:gaussian_bins}
  \centering
  \begin{tabular}{ S S }
  \toprule
  {Mean} & {Variance} \\
  \midrule
  -1.645 &      .080 \\
  -1.080 &      .029 \\
  -.739 &       .018 \\
  -.468 &       .014 \\
  -.228 &       .013 \\
  .000 &        .013 \\
  .228 &        .013 \\
  .468 &        .014 \\
  .739 &        .018 \\
  1.080 &       .029 \\
  1.645 &       .080 \\
  \bottomrule
  \end{tabular}
\end{table*}

\begin{figure}[H]
  \centering
  \includegraphics[width=0.9\linewidth]{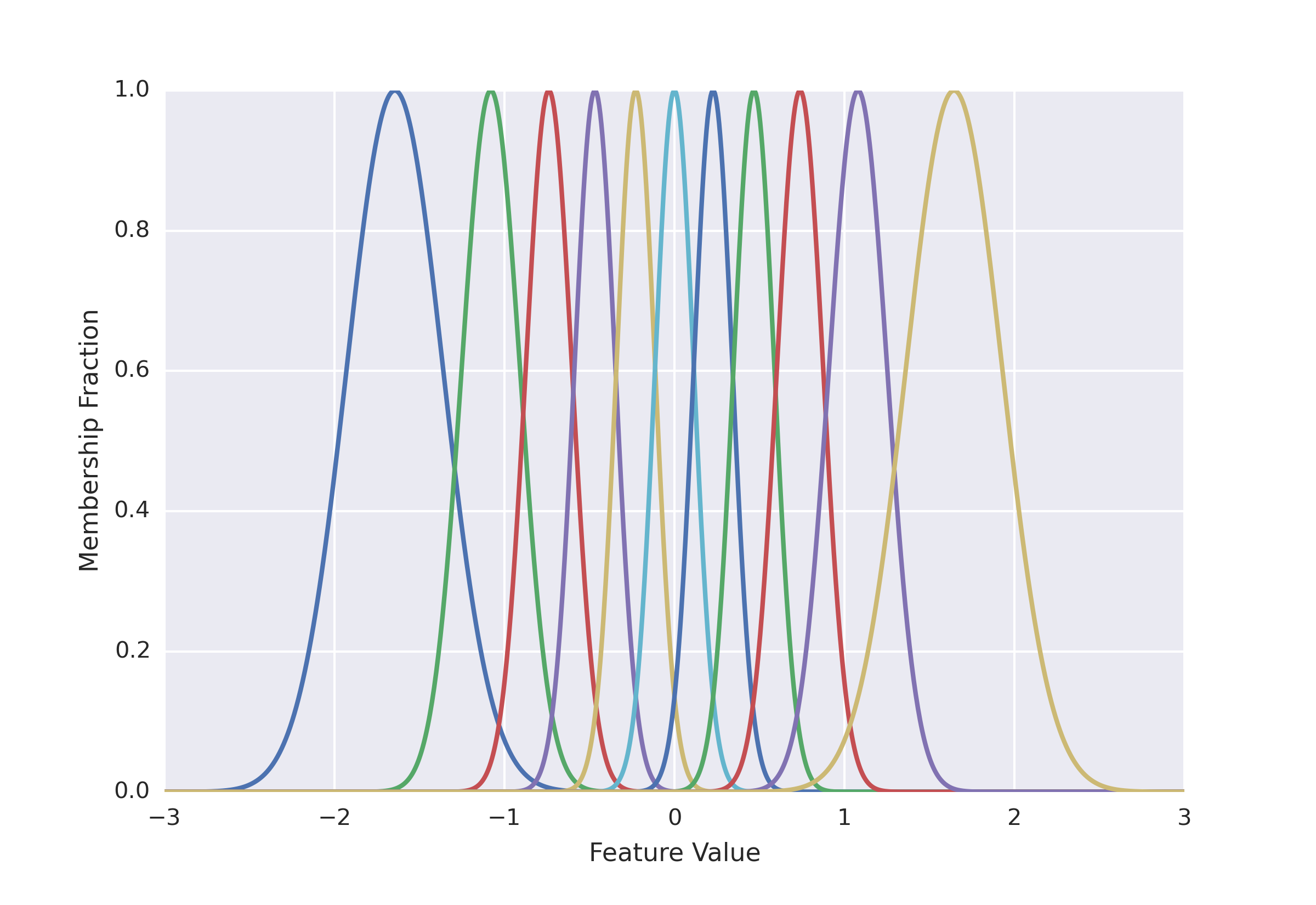}
  \caption{
    Visualization of the Gaussian membership functions.
  }
  \label{appendix:fig:gaussian_bins}
\end{figure}

\putbib
\end{bibunit}

\end{document}